\documentclass{article}

\usepackage{PRIMEarxiv}

\usepackage[utf8]{inputenc}
\usepackage[T1]{fontenc}
\usepackage{hyperref}
\usepackage{url}
\usepackage{booktabs}
\usepackage{amsfonts}
\usepackage{nicefrac}
\usepackage{microtype}
\usepackage{lipsum}
\usepackage{fancyhdr}
\usepackage{graphicx}
\graphicspath{{media/}}

\usepackage{amsmath,amssymb}
\usepackage{multirow}
\usepackage{diagbox}
\usepackage{subcaption}
\usepackage{float}
\usepackage{colortbl}
\usepackage[table]{xcolor}
\usepackage{enumitem}
\usepackage{tcolorbox}
\tcbuselibrary{listings,breakable,skins}
\usepackage{orcidlink}

\title{CAMEO: A Conditional and Quality-Aware Multi-Agent Image Editing Orchestrator}

\author{
\textbf{Yuhan Pu}$^{1}$ \hspace{1.2em}
\textbf{Hao Zheng}$^{2}$ \hspace{1.2em}
\textbf{Ziqian Mo}$^{3}$ \hspace{1.2em}
\textbf{Zirui Pang}$^{1}$ \\[0.4em]
\textbf{Yang Zhang}$^{4}$ \hspace{1.2em}
\textbf{Tianyi Fan}$^{5}$ \hspace{1.2em}
\textbf{Shuhong Wu}$^{5}$ \hspace{1.2em}
\textbf{Jiaheng Wei}$^{1*}$
}

\date{}

\begin{document}
\maketitle
\vspace{-4em}
\begin{center}
{\small
$^{1}$ The Hong Kong University of Science and Technology (Guangzhou), Guangzhou, Guangdong, China \\
$^{2}$ Harbin Institute of Technology, Weihai, Shandong, China \\
$^{3}$ Shenzhen University, Shenzhen, Guangdong, China \\
$^{4}$ Claremont McKenna College, Claremont, California, USA \\
$^{5}$ Research Institute of Petroleum Exploration and Development, CNPC, Beijing, China \\
$^{*}$ Corresponding author: \texttt{jiahengwei@hkust-gz.edu.cn}
}
\end{center}
\vspace{2em}

\begin{abstract}
Conditional image editing aims to modify a source image according to textual prompts and optional reference guidance. Such editing is crucial in scenarios requiring strict structural control (i.e., anomaly insertion in driving scenes and complex human pose transformation). Despite recent advances in large-scale editing models (i.e., Seedream, Nano Banana, etc), most approaches rely on single-step generation. This paradigm often lacks explicit quality control, may introduce excessive deviation from the original image, and frequently produces structural artifacts or environment-inconsistent modifications, typically requiring manual prompt tuning to achieve acceptable results. We propose \textbf{CAMEO}, a structured multi-agent framework that reformulates conditional editing as a quality-aware, feedback-driven process rather than a one-shot generation task. CAMEO decomposes editing into coordinated stages of planning, structured prompting, hypothesis generation, and adaptive reference grounding, where external guidance is invoked only when task complexity requires it. To overcome the lack of intrinsic quality control in existing methods, evaluation is embedded directly within the editing loop. Intermediate results are iteratively refined through structured feedback, forming a closed-loop process that progressively corrects structural and contextual inconsistencies. We evaluate CAMEO on anomaly insertion and human pose switching tasks. Across multiple strong editing backbones and independent evaluation models, CAMEO consistently achieves 20\% more win rate on average compared to multiple state-of-the-art models, demonstrating improved robustness, controllability, and structural reliability in conditional image editing.
 
 % \keywords{Conditional Image Editing \and  Multi-Agent Framework \and Quality Aware}
\end{abstract}

\section{Introduction}

Conditional image editing modifies a source image according to textual instructions, optionally with reference guidance.
Recent advances in diffusion-based generative models have significantly improved semantic alignment and visual realism \cite{rombach2022high, lee2024microstructure, huang2021variational}. Instruction-guided editing systems such as InstructPix2Pix \cite{brooks2023instructpix2pix}, attention-based control methods including Prompt-to-Prompt \cite{hertz2022prompt}, and mask-guided approaches such as DiffEdit \cite{couairon2022diffedit} enable localized modifications while preserving global structure.
Conditional control mechanisms like ControlNet \cite{zhang2023adding} and T2I-Adapter \cite{mou2024t2i} further enhance structural controllability through auxiliary signals. These developments have broadened the use of image editing in content creation, data augmentation, simulation environments, and human-centered applications.

\noindent Despite this progress, conditional image editing remains difficult when multiple structural and contextual constraints must be satisfied simultaneously \cite{perarnau2016invertible, huang2025diffusion}.
In many real-world applications, inserted objects should remain physically plausible, structural transformations should preserve geometric and anatomical consistency, and edits should remain contextually coherent with the surrounding scene.
Such requirements are especially important in safety-sensitive scenarios such as synthetic data generation for autonomous driving, realistic scene simulation, and complex pose manipulation \cite{ma2017pose, yu2020bdd100k, shen2021igibson}. As editing complexity increases, maintaining these constraints consistently becomes substantially more challenging.

\noindent\textbf{Open-loop generation under multi-constraint settings.}
Most existing editing systems still rely on single-pass generation \cite{rombach2022high, brooks2023instructpix2pix}, where semantic alignment, geometric consistency, physical plausibility, and contextual coherence must be satisfied within a single generation step. In practice, this often leads to distorted structures, inconsistent illumination, or scene-inconsistent edits when transformations become large or structurally demanding.

\noindent\textbf{Lack of intrinsic quality control.}
Although evaluation metrics such as CLIP similarity \cite{radford2021learning} and perceptual measures such as LPIPS \cite{zhang2018unreasonable, karnatov2025analysis} provide useful post hoc assessment, generation and evaluation remain largely decoupled. Correcting errors often requires repeated sampling or manual prompt tuning, limiting robustness and scalability in multi-constraint scenarios, resulting in plenty of label noise \cite{wei2021learning} while implementing benchmarks.

\noindent\textbf{Rigid reference conditioning.}
Structural guidance signals, including pose maps, segmentation maps, and reference images \cite{zhang2023adding, mou2024t2i, titov2024guide}, are typically applied uniformly rather than adaptively.
While such conditioning improves controllability, it does not dynamically adjust to varying task difficulty or transformation magnitude.

\begin{figure}[H]
\centering
\includegraphics[width=\linewidth]{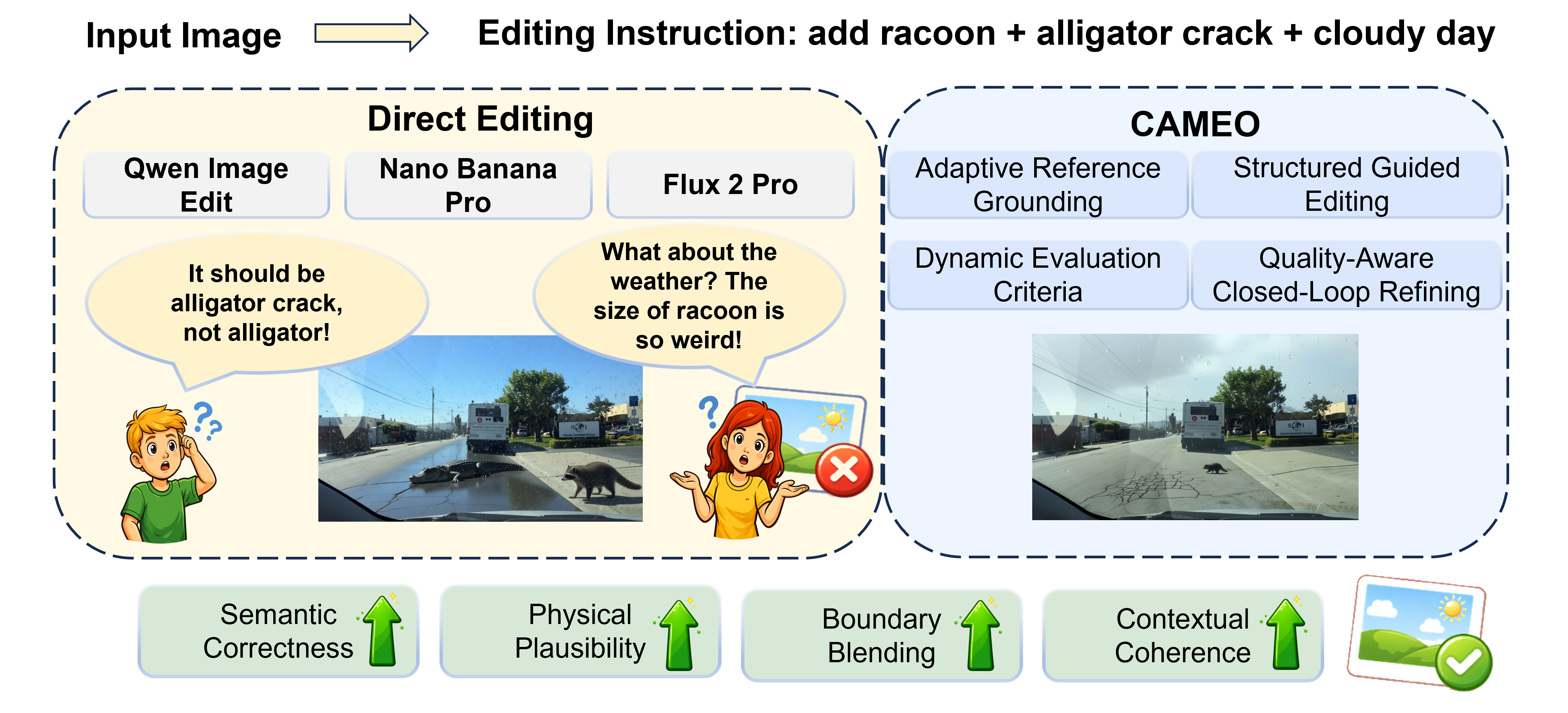}
\caption{Compare CAMEO with other State-of-the-Art image editing models}
\label{fig:general_image}
\end{figure}

\noindent To address these limitations, we propose \textbf{CAMEO}, a hierarchical multi-agent framework that treats conditional editing as an iterative process with embedded evaluation and feedback.
Rather than relying solely on a single-pass transformation, CAMEO organizes editing into coordinated stages of task interpretation, prompt construction, adaptive reference grounding, structured generation, quality evaluation on dynamic criterion, and iterative refinement. This design progressively reduces structural and contextual inconsistencies while reducing dependence on repeated unguided sampling. An example comparison between CAMEO and representative image editing models is presented in Fig.~\ref{fig:general_image}.

\noindent We evaluate CAMEO on two challenging tasks: road anomaly insertion on BDD100K \cite{yu2020bdd100k} and human pose switching. Across multiple editing backbones and independent vision-language judges, CAMEO consistently outperforms direct editing baselines. Human preference studies further confirm improvements in semantic correctness, physical plausibility, boundary blending and contextual coherence.

\noindent Our contributions are fourfold:
\begin{itemize}[leftmargin=1.0em,label=\textbullet]

\item We introduce \textbf{CAMEO}, a hierarchical multi-agent architecture that decomposes conditional image editing into orchestration, execution, and regulation tiers (\S3). This structured design replaces monolithic single-pass pipelines with coordinated functional roles tailored for complex editing scenarios.

\item We reformulate multi-constraint conditional image editing as an explicitly regulated optimization process rather than implicit constraint satisfaction within a single generative trajectory (\S3.2–\S3.4). This paradigm shift enables progressive constraint verification and controlled correction during synthesis.

\item We construct a dedicated benchmark for complex \textbf{human pose switching}, explicitly designed to evaluate structural validity, physical plausibility, and contextual coherence under large pose transformations (\S4.6). This benchmark complements existing editing datasets by introducing multi-constraint evaluation settings for articulated human motion.

\item We conduct extensive experiments across multiple editing backbones and independent vision-language judges, complemented by human evaluation, demonstrate consistent gains in robustness and controllability over direct editing baselines (\S4).

\end{itemize}

\begin{figure}[H]
\centering
\includegraphics[width=0.48\linewidth,height=0.28\linewidth]{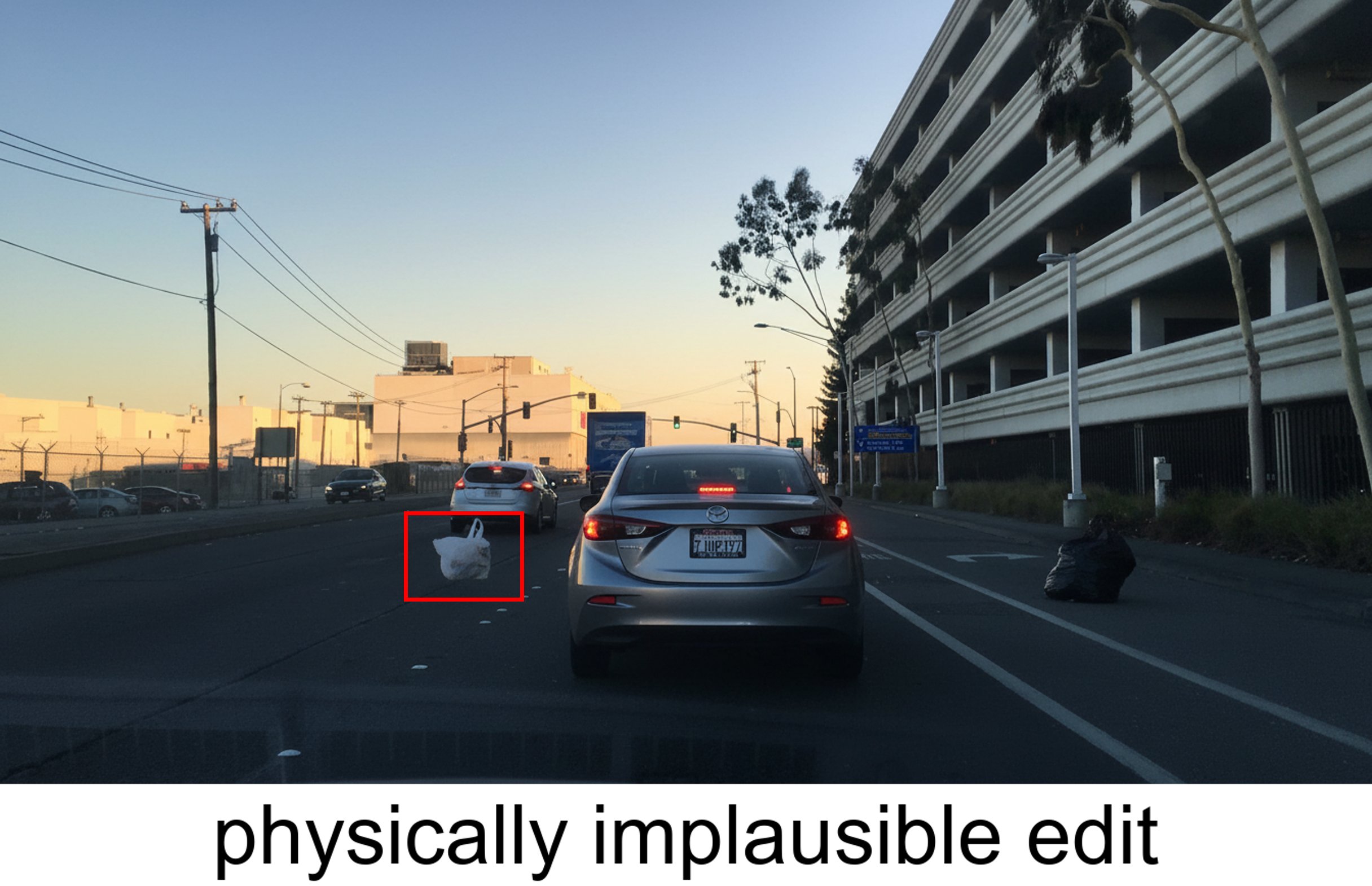}
\hfill
\includegraphics[width=0.48\linewidth,height=0.28\linewidth]{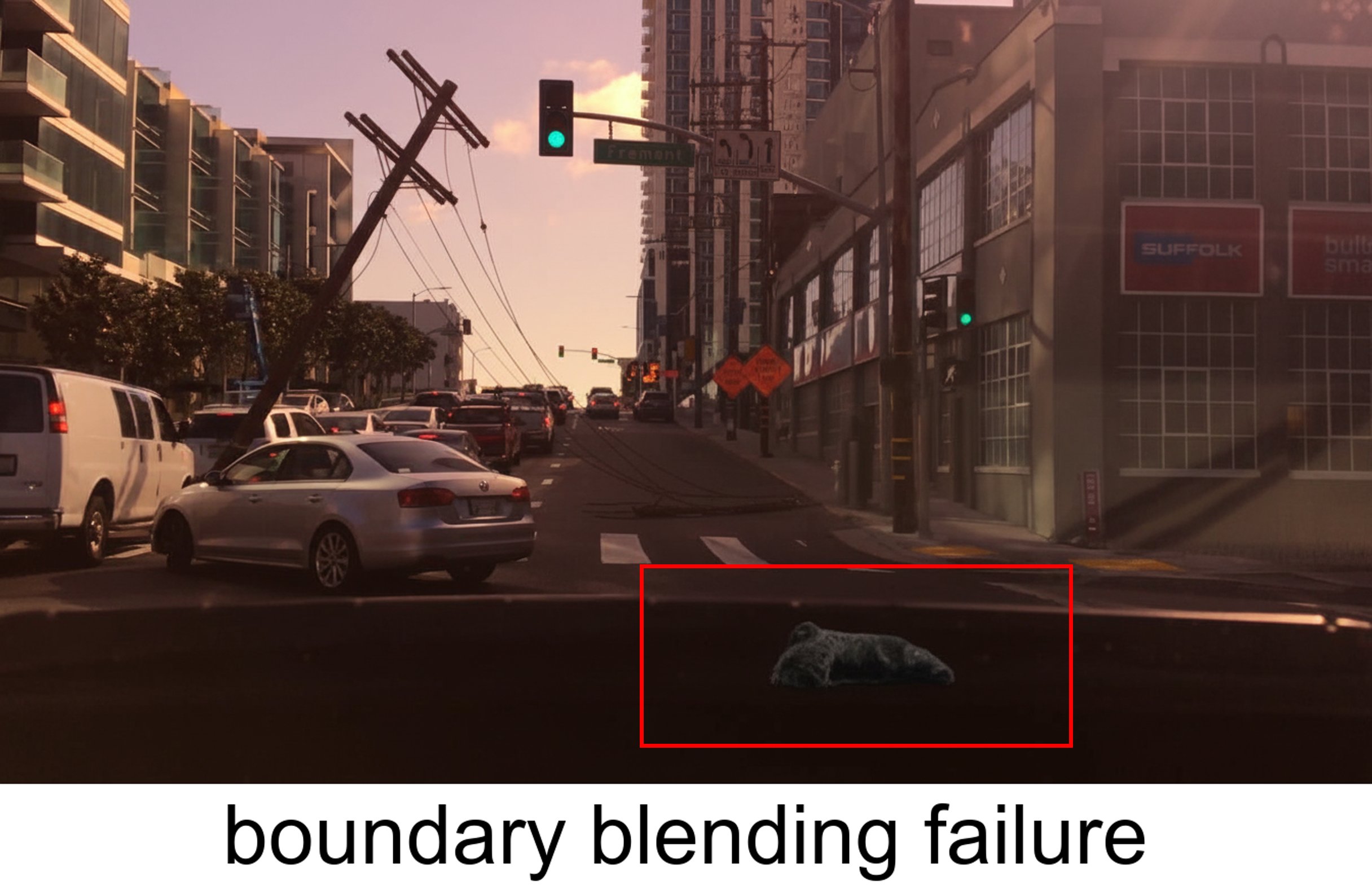}

\includegraphics[width=0.48\linewidth,height=0.28\linewidth]{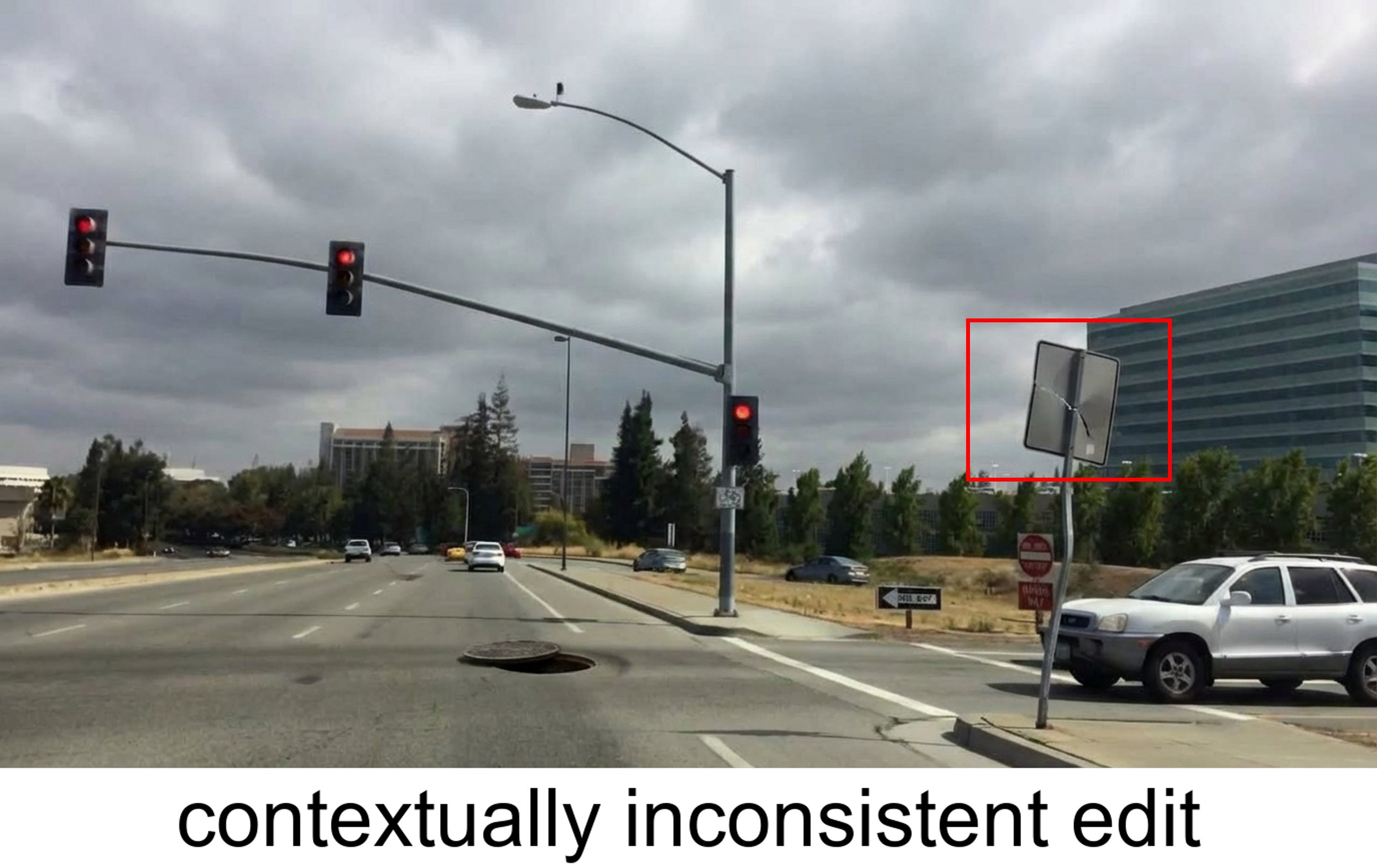}
\hfill
\includegraphics[width=0.48\linewidth,height=0.28\linewidth]{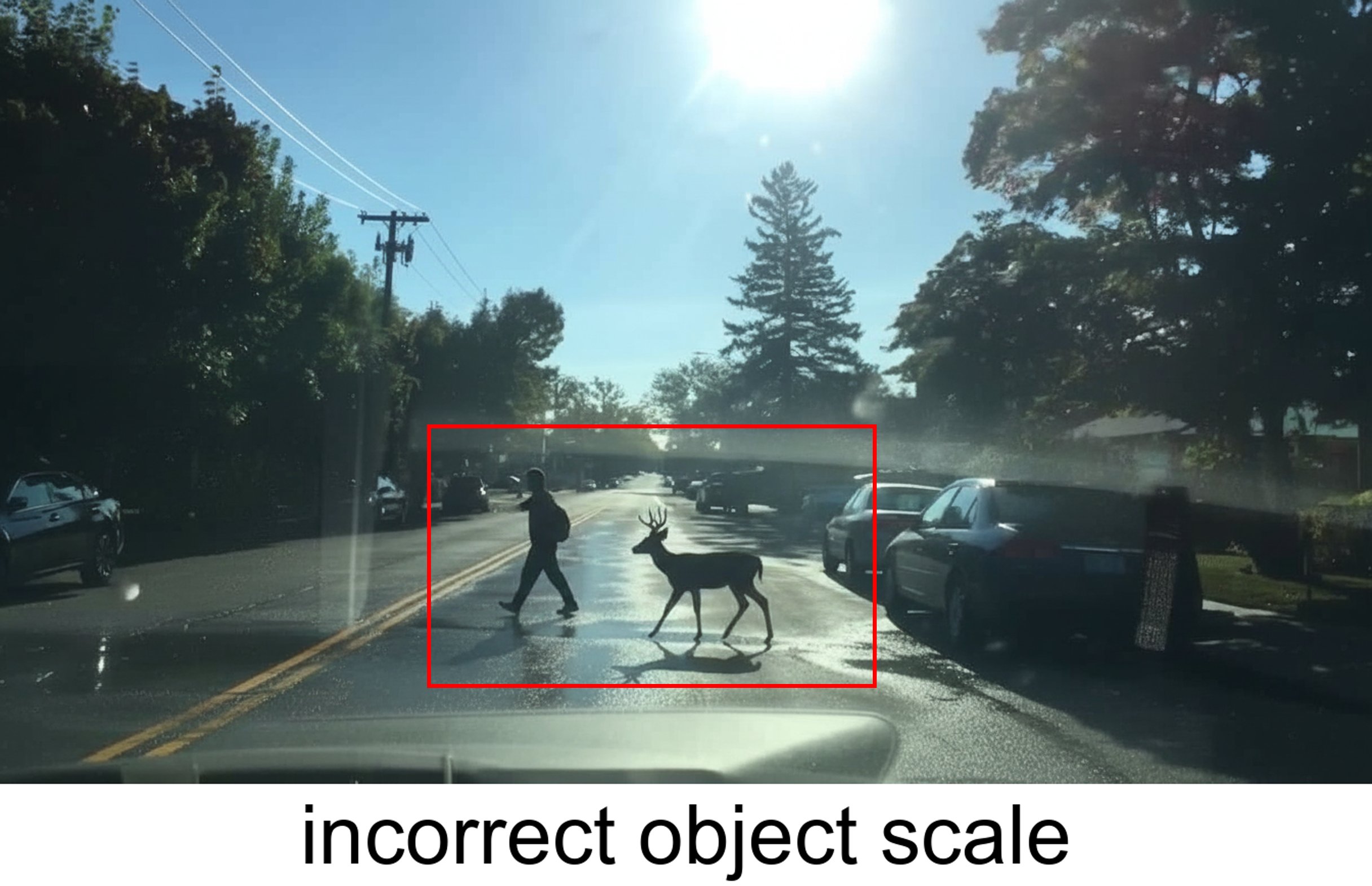}

\caption{
Representative failure cases illustrating common issues of conditional image editing on images from BDD100K under complex situations. 
}
\label{fig:badcases}
\end{figure}

\section{Related Work}

\noindent\textbf{State-of-the-Art Image Editing Models.}
Recent years have witnessed rapid progress in instruction-based image editing, driven by diffusion models and multimodal large language models. InstructPix2Pix shows that editing can be learned from synthetic instruction-image pairs within a diffusion framework \cite{brooks2023instructpix2pix}, while latent diffusion improves high-resolution synthesis in a compressed latent space \cite{rombach2022high}. Subsequent works further enhance controllability and semantic alignment, including Prompt-to-Prompt \cite{hertz2022prompt}, DiffEdit \cite{couairon2022diffedit}, Imagic \cite{kawar2023imagic}, Plug-and-Play Diffusion Features \cite{tumanyan2023plug}, and ControlNet \cite{zhang2023sine}. More recent approaches explore richer instruction interfaces and multimodal reasoning, such as MGIE \cite{fu2023guiding} and GenArtist \cite{wang2024genartist}, while subject-driven and compositional editing are studied in DreamBooth \cite{ruiz2023dreambooth}, Blended Diffusion \cite{avrahami2022blended}, SDEdit \cite{meng2021sdedit}, and image translation methods such as Detail Fusion GAN \cite{li2021detail}. Commercial systems such as Qwen Image Edit Plus, FLUX 2 Pro, Seedream 4.5, and Nano Banana Pro further demonstrate strong progress in controllability and fidelity. At the same time, recent studies reveal that modern multimodal and editing systems remain vulnerable to robustness, safety, and misinformation-related issues, highlighting the need for stronger control and verification mechanisms \cite{cheng2025tvpi, chen2025safeeraser, zheng2025offside}. Despite these advances, most existing methods still rely on single-pass generation and lack explicit decomposition and iterative verification for complex multi-constraint editing tasks.

\noindent\textbf{Image Editing Evaluations.}
Evaluating image editing remains difficult because there is usually no unique ground-truth target and editing quality is inherently multi-dimensional. Traditional metrics such as FID measure distribution-level realism \cite{heusel2017gans}, LPIPS evaluates perceptual similarity \cite{zhang2018unreasonable}, and CLIPScore measures text-image alignment without paired references \cite{hessel2021clipscore}. Related multimodal evaluation methods also use pretrained vision-language models such as CLIP to assess semantic consistency \cite{radford2021learning}. Beyond editing, recent studies have also explored the use of VLMs in fine-grained recognition, missing-label discovery, and visually complex classification settings, suggesting their broader potential for nuanced visual assessment \cite{pang2025vlms, ma2021research}. However, these metrics do not directly capture structural correctness, physical plausibility, or contextual coherence in edited regions. Human preference learning and pairwise comparison protocols provide complementary evaluation perspectives \cite{christiano2017deep, ouyang2022training}. These limitations are especially clear in human pose editing and transfer, where strict anatomical and geometric constraints are required. Earlier methods such as PG$^2$ \cite{ma2017pose}, deformable GAN-based models \cite{siarohin2018deformable}, and later correspondence- or motion-based approaches \cite{siarohin2019first, chan2019everybody} establish important settings for pose manipulation, but their evaluations mainly emphasize pose alignment or perceptual similarity. Related vision tasks such as semantic tracking and geo-localization also emphasize precise spatial correspondence and fine-grained structure, which further highlights the importance of constraint-aware evaluation \cite{wang2024semtrack, mo2025sign}. This motivates more constraint-aware evaluation protocols for complex conditional image editing.

\noindent\textbf{Image Editing Benchmarks.}
To overcome the limitations of generic metrics, recent works have introduced dedicated editing benchmarks. I$^2$EBench evaluates perceptual, semantic, and structural aspects of instruction-guided editing \cite{ma2024i2ebench}, while LMM4Edit scales evaluation with large-scale human preference annotations \cite{xu2025lmm4edit}. More recent benchmarks target increasingly realistic and challenging settings. KRIS-Bench emphasizes knowledge-based reasoning in editing \cite{wu2025krisbench}; CompBench studies fine-grained instruction following, spatial reasoning, and contextual reasoning \cite{jia2025compbench}; RefEdit-Bench focuses on referring-expression-based editing in complex multi-entity scenes \cite{pathiraja2025refedit}; and ImgEdit-Bench provides a unified benchmark covering instruction adherence, editing quality, detail preservation, and both single-turn and multi-turn settings \cite{ye2025imgedit}. FragFake further examines fine-grained edited-image detection and localization, reflecting growing interest in both editing quality and manipulation authenticity \cite{sun2025fragfake}. Related benchmark construction efforts in adjacent visual domains, including tracking, multimodal unlearning, and remote-sensing perception, also reflect the broader trend toward more challenging and structured evaluation settings \cite{wang2024semtrack, zheng2025offside, xu2025enhanced}.

\noindent\textbf{Structured and Multi-Agent Frameworks.}
Structured generation has become an important paradigm for handling tasks with multiple interdependent constraints. Instead of treating generation as a single process, modular frameworks decompose it into specialized components, improving interpretability and controllability \cite{yao2022react, shinn2023reflexion, schick2023toolformer, wu2023visual, park2023generative}. This idea has also been extended to image generation and editing. GenArtist frames image generation and editing as an agent-driven process \cite{wang2024genartist}, ComfyMind explores tree-based planning and reactive feedback \cite{guo2025comfymind}, and MIRA formulates editing as an iterative perception-reasoning-action loop \cite{zeng2025mira}. Related work in spatial modeling, multimodal safety, and structured visual reasoning also suggests the value of decomposing complex visual tasks into coordinated modules \cite{chen2025safeeraser, mo2025sign, xu2025enhanced}. These works suggest that decomposition and iterative reasoning can improve robustness over single-pass generation. However, existing systems are often application-specific or loosely coupled, and usually do not unify task planning, reference retrieval, generation, quality assessment, and iterative refinement within one framework. Recent multi-agent editing studies \cite{venkatesh2025crea, ma2025talk2image, fu2024precise, ye2026agent} further support this direction, but adaptive reference grounding and multi-dimensional evaluation remain underexplored.

\section{Method}
This section first introduces the overall architectural design and agent hierarchy, followed by detailed descriptions of task coordination, constraint regulation, and iterative refinement mechanisms. Together, these components enable explicit constraint management under complex multi-constraint editing scenarios.
\subsection{Hierarchical Agent Architecture}

CAMEO is built on a hierarchical multi-agent architecture designed to decompose conditional image editing into coordinated and controllable components. Rather than treating editing as a monolithic generation procedure, CAMEO distributes responsibilities across specialized agents organized into three functional tiers: orchestration agents, utility agents, and regulation agents.

\noindent\textbf{Orchestration Agents.}
At the top tier, the \textit{Strategic Director} serves as the global controller. It interprets task intent, determines the active constraint dimensions, and decides whether additional reference grounding is required. By dynamically allocating responsibilities and activating constraint sets, the Strategic Director adapts the editing process to task complexity and transformation difficulty.

\noindent\textbf{Utility Agents.}
The middle tier consists of constructive agents responsible for generating candidate solutions.
The \textit{Instruction Architect} converts high-level instructions into structured, constraint-enriched prompts.
The \textit{Visual Research Specialist} provides adaptive guidance by retrieving or synthesizing textual and/or visual references. Depending on task requirements, the Visual Research Specialist may operate in textual mode, visual mode, hybrid mode, or remain inactive. The \textit{Generative Creator} interfaces with backbone editing models to produce candidate hypotheses conditioned on structured prompts and optional references.

\noindent\textbf{Regulation Agents.}
The bottom tier enforces intrinsic quality control.
The \textit{Quality Critic} evaluates intermediate results under task-adaptive criteria and produces structured diagnostic feedback. The \textit{Refinement Editor} applies targeted corrections guided by this feedback, progressively reducing structural and contextual inconsistencies. This tier transforms editing from an open-loop generation process into a self-regulating system.

\begin{figure}[H]
\centering
\includegraphics[width=0.9\linewidth]{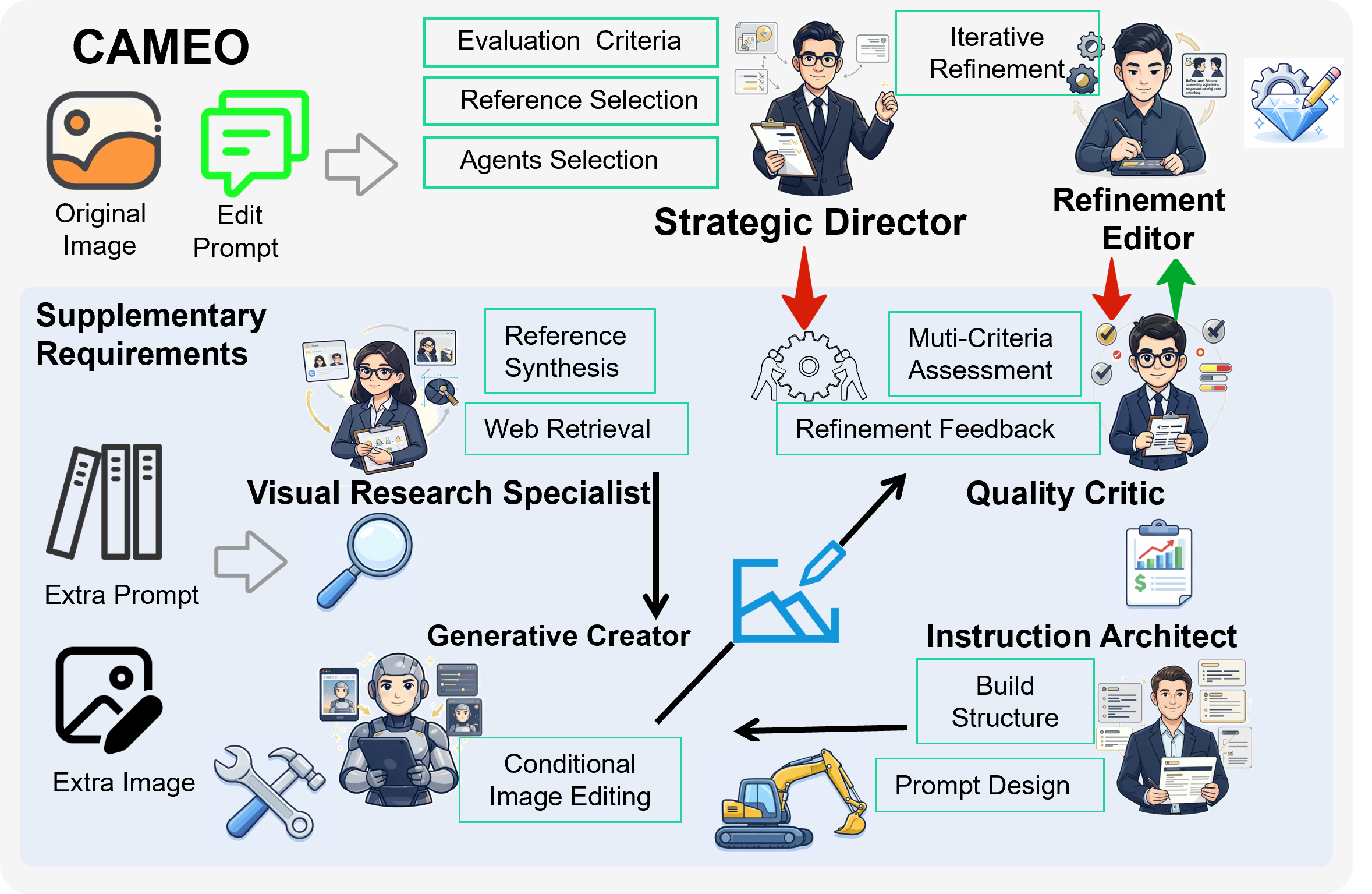}
\caption{Overview of the CAMEO multi-agent workflow. The Strategic Director coordinates multiple agents to perform task interpretation, structured generation, quality evaluation, and iterative refinement. }
\label{fig:agent_workflow}
\end{figure}

\noindent The specific models employed by each agent and the rationale for their selection are provided in the Appendix. Through hierarchical specialization, CAMEO balances flexibility, controllability, and robustness across diverse conditional editing tasks. An overview of the proposed agent-based workflow is illustrated in Fig.~\ref{fig:agent_workflow}.

\subsection{Overall Workflow}

Given a source image $I$ and instruction $T$, CAMEO produces an edited image $\hat{I}$ through coordinated multi-agent interaction. The workflow proceeds in four stages:
\begin{itemize}[leftmargin=1.0em,label=\textbullet]
\item \textbf{Task Interpretation.}
The Strategic Director analyzes $I$ and $T$, selects task-adaptive evaluation criteria, and determines the necessity and type of reference grounding.

\item \textbf{Structured Generation.}
The Instruction Architect constructs constraint-enriched prompts.
If required, the Visual Research Specialist supplements the editing context with textual and/or visual priors.
The Generative Creator produces an initial hypothesis $\tilde{I}^{(0)}$.

\item \textbf{Quality Evaluation.}
The Quality Critic evaluates $\tilde{I}^{(t)}$ under selected constraints and produces diagnostic signals.

\item \textbf{Iterative Refinement.}
The Refinement Editor updates the hypothesis based on structured feedback until quality thresholds are satisfied.

\end{itemize}

\noindent Unlike conventional single-pass editing pipelines, CAMEO explicitly embeds evaluation and correction within the generation loop, enabling progressive constraint enforcement.

\subsection{Adaptive Reference Grounding}

External guidance can significantly improve structural fidelity, but excessive conditioning may introduce bias or over-constrain the editing trajectory. CAMEO therefore adopts adaptive reference grounding. Let the reference configuration be defined as
\vspace{1em}
\[
\mathcal{R} \in \{\varnothing, \mathcal{R}_T, \mathcal{R}_V, \mathcal{R}_{TV}\},
\]
\vspace{1em}
where $\varnothing$ denotes no reference, $\mathcal{R}_T$ denotes textual references, $\mathcal{R}_V$ denotes visual references, and $\mathcal{R}_{TV}$ denotes hybrid textual–visual references.

%The Visual Research Specialist dynamically selects one of the following modes:

%\begin{itemize}[leftmargin=1.0em,label=\textbullet]

%\item \textbf{No reference:}
%\[
%\mathcal{R} = \varnothing.
%\]

%\item \textbf{Textual reference only:}
%\[
%\mathcal{R} = \mathcal{R}_T.
%\]

%\item \textbf{Visual reference only:}
%\[
%\mathcal{R} = \mathcal{R}_V.
%\]

%\item \textbf{Hybrid textual–visual reference:}
%\[
%\mathcal{R} = \mathcal{R}_{TV} = \{\mathcal{R}_T, \mathcal{R}_V\}.
%\]

%\end{itemize}

\noindent The reference mode is dynamically selected by the Visual Research Specialist under the guidance of the Strategic Director based on task complexity and constraint sensitivity, ensuring that reference signals are introduced only when necessary.

\subsection{Quality-Aware Closed-Loop Editing}

A core design principle of CAMEO is intrinsic quality control.

\noindent Given an intermediate hypothesis $\tilde{I}^{(t)}$, the Quality Critic evaluates constraint satisfaction:

\[
\Delta^{(t)} = \mathcal{A}_{qa}(\tilde{I}^{(t)}, I, T, \mathcal{R}),
\]

where $\Delta^{(t)}$ denotes a structured constraint deviation signal encoding semantic, structural, and contextual violations at iteration $t$.

\noindent The Refinement Editor updates the hypothesis:
\vspace{0.5em}
\[
\tilde{I}^{(t+1)} = \mathcal{R}_{edit}(\tilde{I}^{(t)}, \Delta^{(t)}).
\]
\vspace{0.3em}
where $\mathcal{R}_{edit}$ denotes the refinement operator parameterized by the underlying editing backbone.
Editing terminates when all active constraints satisfy predefined thresholds or when iteration limits are reached.

\noindent By embedding evaluation within the generation loop, CAMEO converts conditional editing into a closed-loop process that progressively improves structural alignment and contextual coherence.

\subsection{A Control-Theoretic Perspective on Conditional Editing}

From a broader perspective, CAMEO can be viewed as introducing structured control into conditional image editing.

\noindent Conventional editing systems operate as open-loop mappings:
\[
\hat{I} = f_\theta(I, T),
\]
where constraint satisfaction is expected to emerge from a single forward pass.
Such formulations provide limited guarantees on structural fidelity or contextual coherence, especially when multiple heterogeneous constraints must be jointly satisfied.

\noindent CAMEO instead introduces an internal control mechanism that continuously monitors and regulates the editing trajectory.
Let $\mathcal{S}^{(t)}$ denote the editing state at iteration $t$.
Quality assessment produces a structured constraint deviation signal $\Delta^{(t)}$, which guides corrective updates:
\[
\mathcal{S}^{(t+1)} = \Phi(\mathcal{S}^{(t)}, \Delta^{(t)}).
\]

where $\Phi$ denotes the closed-loop state transition function induced by refinement and regulation.

\noindent Under this formulation, editing becomes a closed-loop control process rather than a single-pass transformation.
The Strategic Director defines the active constraints, the Visual Research Specialist adjusts structural priors, and the Refinement Editor reduces deviation signals over time.
This hierarchical coordination progressively stabilizes structural alignment and contextual coherence. In this sense, CAMEO transforms conditional editing from implicit constraint satisfaction into explicit constraint regulation through multi-agent interaction.

\begin{figure}[!htbp]
    \centering
    
    \begin{subfigure}{\linewidth}
        \centering
        \includegraphics[width=\linewidth]{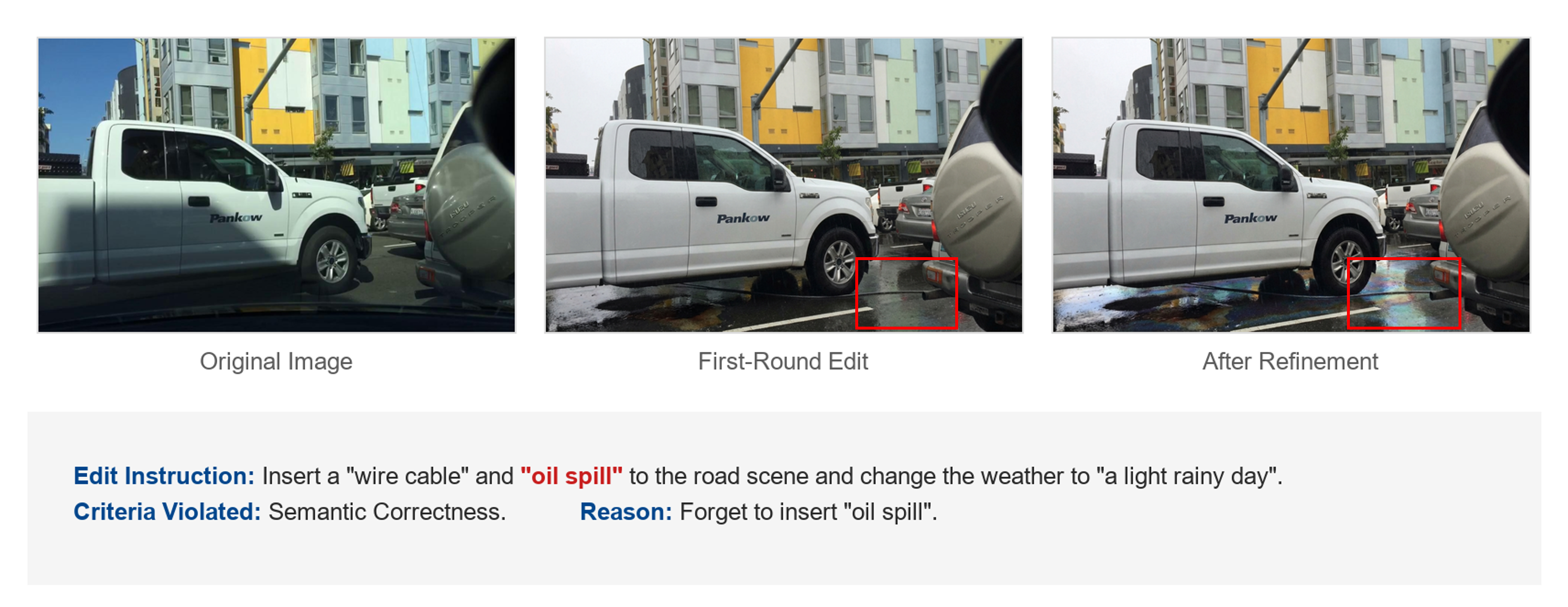}
    \end{subfigure}    
    \begin{subfigure}{\linewidth}
        \centering
        \includegraphics[width=\linewidth]{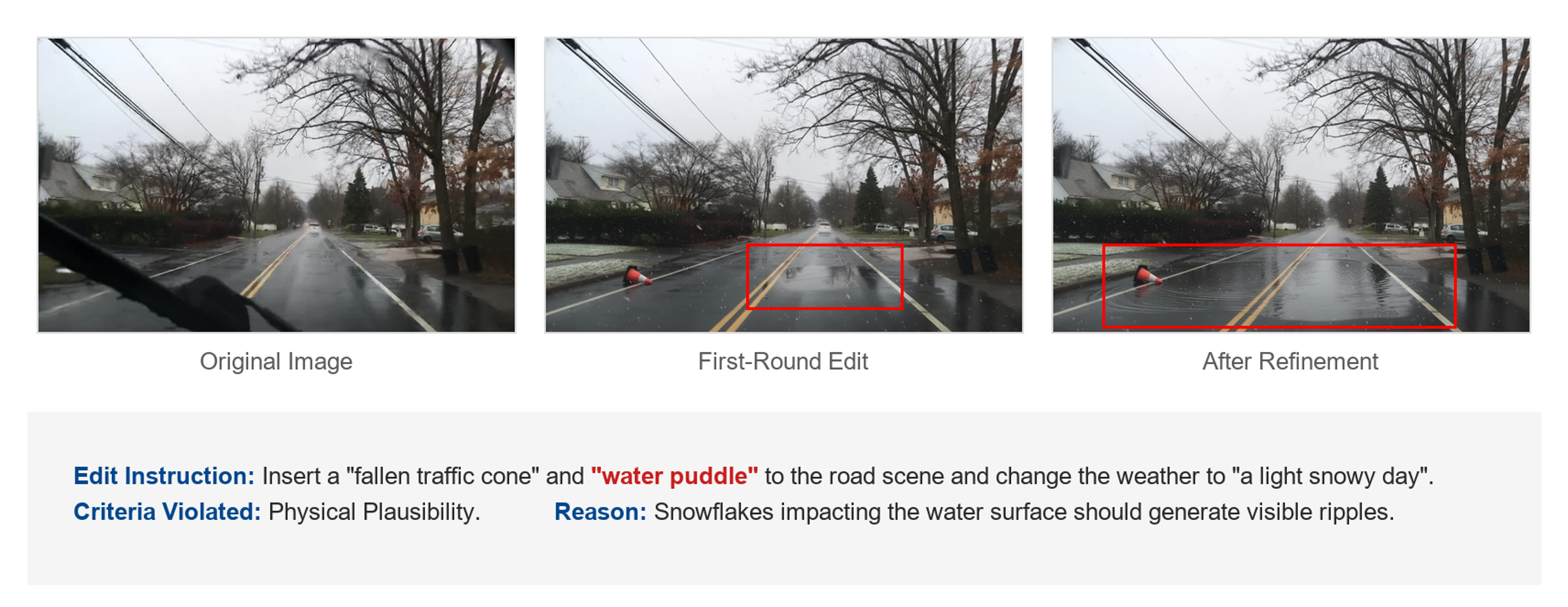}
    \end{subfigure}
    
    \caption{Representative cases of how CAMEO improves semantic correctness and physical plausibility issues.}
    \label{fig:group1}
\end{figure}
\begin{figure}[!htbp]
    \centering
    
    \begin{subfigure}{\linewidth}
        \centering
        \includegraphics[width=\linewidth]{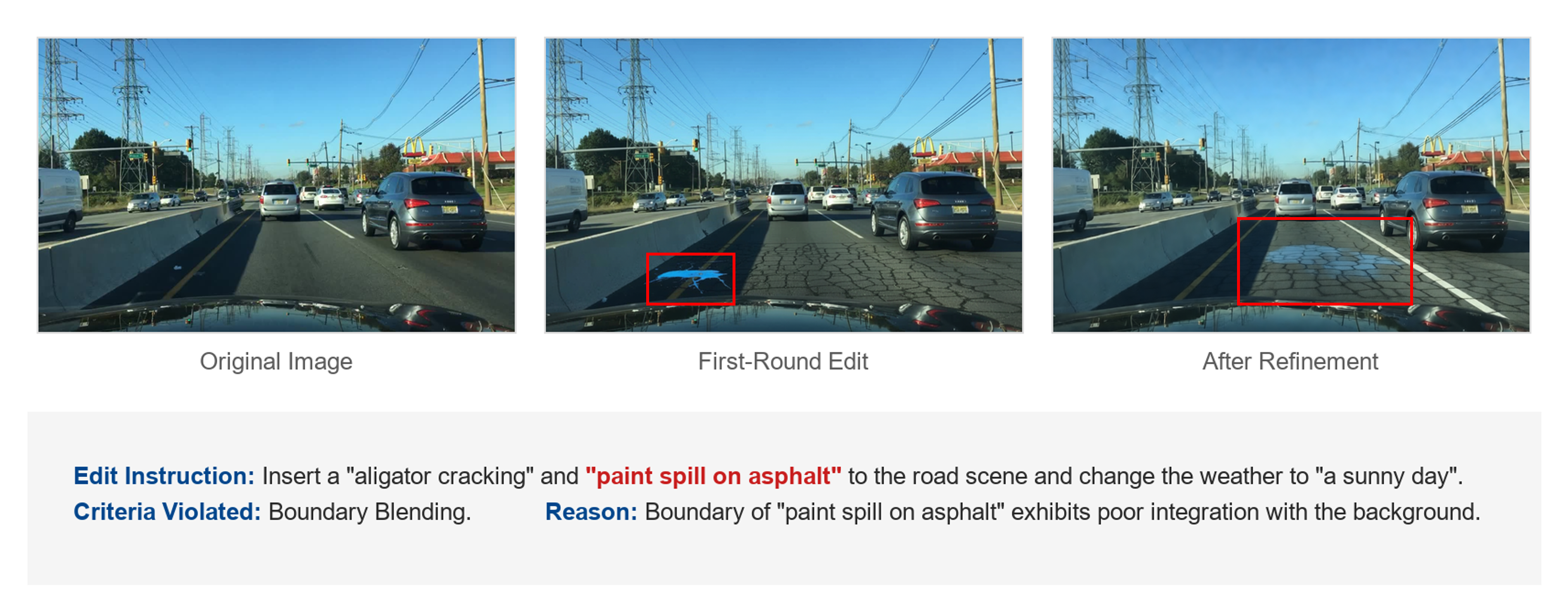}
    \end{subfigure}    
    \begin{subfigure}{\linewidth}
        \centering
        \includegraphics[width=\linewidth]{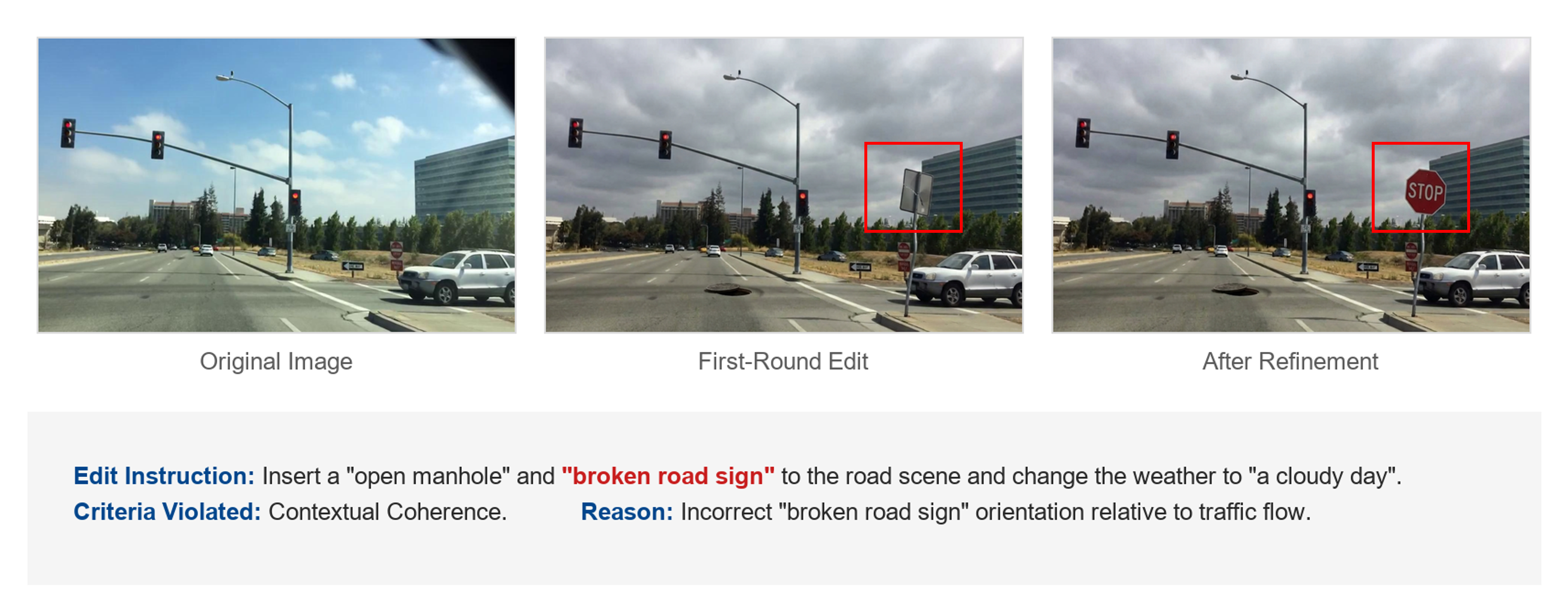}
    \end{subfigure}
    
    \caption{Representative cases of how CAMEO improves boundary blending and contextual coherence issues.}
    \label{fig:group2}
\end{figure}

\section{Experiments}
\subsection{Experimental Setup}

We evaluate CAMEO on two representative conditional editing tasks: road anomaly insertion and human pose switching.

\noindent\textbf{Backbone Models.}
CAMEO is implemented on top of multiple strong editing backbones, including Qwen Image Edit Plus, FLUX 2 Pro, Seedream 4.5, and Nano Banana Pro.
For each backbone, we compare direct single-step editing against CAMEO-enhanced editing under identical inputs.

\noindent\textbf{Evaluation Protocol.}
We adopt an arena-style pairwise comparison protocol.
For each test case, baseline and CAMEO outputs are evaluated by multiple independent vision-language judges (Qwen3-VL-Plus, GPT-4o, Gemini-2.5, and Claude-Opus-4.5).
Each judge provides (1) a win/lose/tie decision and (2) a comprehensive score from 1 to 10 for each image.  To mitigate position bias, image order is alternated across evaluation rounds, ensuring each method to appear first with equal probability. For detailed evaluation criteria and prompts, please refer to the Appendix. We report aggregated win rates and average score differences.

\subsection{Road Anomaly Insertion}

We conduct a large-scale evaluation using 10,000 images sampled from the BDD100K dataset. For each image, an anomaly insertion instruction is randomly selected from a predefined set of 30 rare road anomaly categories under 10 different weather conditions, ensuring broad coverage and reducing category-specific bias. As shown in Table~\ref{tab:anomaly_results}, CAMEO achieves higher win rates across all backbone models and vision-language judges, with particularly clear advantages in scenarios requiring physical plausibility and contextual coherence. On average, CAMEO achieves about a 20\% higher win rate than direct editing. Table~\ref{tab:avg_scores_anomaly} further reports the average evaluation scores, where CAMEO consistently outperforms baselines across all four judges, indicating improved editing quality under complex road anomaly insertion. Detailed examples illustrating how CAMEO improves image quality are shown in Fig.~\ref{fig:group1} and Fig.~\ref{fig:group2}.

\subsection{Human Pose Switching}

\begin{figure}[!htbp]
    \centering
    \includegraphics[width=\linewidth]{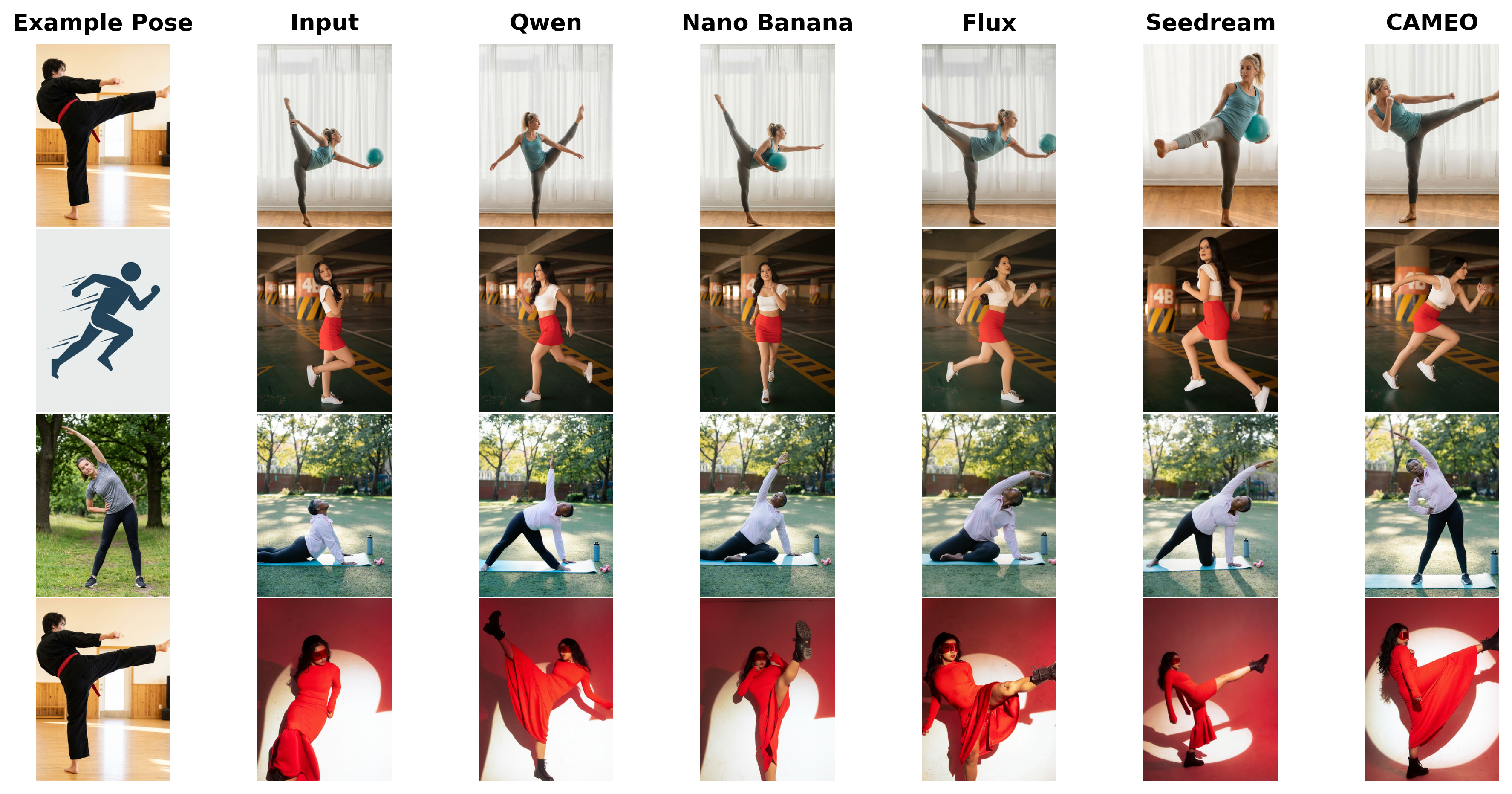}
    \caption{Qualitative comparison across methods on diverse human pose switching examples. Each input consists of an instruction and an original image. Instructions are:
    \textbf{Row 1:}  a high front kick with the right leg extended forward, upper body leaning back for balance.
    \textbf{Row 2:} a running posture with the right leg stepping forward, left leg pushing back. 
    \textbf{Row 3:} a side stretching pose bending the torso to the right with left arm reaching overhead. 
    \textbf{Row 4:} a high front kick with the right leg extended forward, upper body leaning back for balance.}
    \label{fig:comparison_gs}
\end{figure}

To evaluate structural transformation capability, we construct a pose switching benchmark using 1,000 full-body human images collected from Pexels via API, covering diverse identities, body types, backgrounds, and viewpoints. For each image, 10 pose modification instructions are randomly sampled from a predefined set of 30 target poses, resulting in 10,000 edited samples. Evaluation focuses on whether generated poses match the target configuration while preserving anatomical plausibility. As shown in Table~\ref{tab:pose_results}, CAMEO consistently outperforms direct editing baselines across all backbone models and independent vision-language judges, with particularly clear improvements for large pose changes where adaptive reference grounding and iterative refinement reduce limb distortion and structural inconsistencies. Fig.~\ref{fig:comparison_gs} shows qualitative comparisons across different editing methods on diverse human pose switching examples. On average, CAMEO achieves about a 20\% higher win rate than direct editing. Table~\ref{tab:avg_scores_pose} further reports the average evaluation scores, where CAMEO demonstrates competitive or superior performance across most backbone–judge combinations, indicating improved robustness in structurally demanding pose transformations.

\begin{table*}[!htbp]
\centering
\caption{Pairwise win/lose/tie statistics (\%) on the road anomaly insertion task across multiple vision-language judges. Higher win rates indicate stronger performance of CAMEO over direct editing.}
\label{tab:anomaly_results}
{
\normalsize
\setlength{\tabcolsep}{1.5pt}
\begin{tabular}{l|ccc|ccc|ccc|ccc}
\toprule

\diagbox{\textbf{Backbone}}{\textbf{Judge}}
& \multicolumn{3}{c|}{\textbf{Qwen3-VL-Plus}}
& \multicolumn{3}{c|}{\textbf{GPT-4o}}
& \multicolumn{3}{c|}{\textbf{Gemini-2.5}}
& \multicolumn{3}{c}{\textbf{Claude-Opus-4.5}} \\

\cmidrule(lr){2-4}
\cmidrule(lr){5-7}
\cmidrule(lr){8-10}
\cmidrule(lr){11-13}

& Win & Lose & Tie
& Win & Lose & Tie
& Win & Lose & Tie
& Win & Lose & Tie\\

\midrule

\textbf{Qwen-Image-Edit-Plus}
& \cellcolor{gray!15}66.41 & 31.92 & 1.67
& \cellcolor{gray!15}60.83 & 36.48 & 2.69
& \cellcolor{gray!15}66.88 & 29.77 & 3.35
& \cellcolor{gray!15}61.37 & 35.94 & 2.69 \\

\textbf{Nano Banana Pro}
& \cellcolor{gray!15}60.30 & 37.55 & 2.15
& \cellcolor{gray!15}59.74 & 38.96 & 1.30
& \cellcolor{gray!15}68.18 & 28.02 & 3.80
& \cellcolor{gray!15}57.89 & 39.84 & 2.27 \\

\textbf{Flux-2-Pro}
& \cellcolor{gray!15}63.31 & 33.91 & 2.78
& \cellcolor{gray!15}58.72 & 39.41 & 1.87
& \cellcolor{gray!15}60.14 & 36.32 & 3.54
& \cellcolor{gray!15}59.21 & 38.22 & 2.57 \\

\textbf{Seedream-4.5}
& \cellcolor{gray!15}65.92 & 31.01 & 3.07
& \cellcolor{gray!15}59.06 & 38.88 & 2.06
& \cellcolor{gray!15}68.18 & 27.46 & 4.36
& \cellcolor{gray!15}56.12 & 40.33 & 3.55 \\

\bottomrule
\end{tabular}
}
\end{table*}

\begin{table*}[!htbp]
\centering
\caption{Pairwise win/lose/tie statistics (\%) on the human pose switching task across multiple vision-language judges. Higher win rates indicate stronger performance of CAMEO over direct editing.}
\label{tab:pose_results}
{
\normalsize
\setlength{\tabcolsep}{1.5pt}
\begin{tabular}{l|ccc|ccc|ccc|ccc}
\toprule

\diagbox{\textbf{Backbone}}{\textbf{Judge}}
& \multicolumn{3}{c|}{\textbf{Qwen3-VL-Plus}}
& \multicolumn{3}{c|}{\textbf{GPT-4o}}
& \multicolumn{3}{c|}{\textbf{Gemini-2.5}}
& \multicolumn{3}{c}{\textbf{Claude-Opus-4.5}} \\

\cmidrule(lr){2-4}
\cmidrule(lr){5-7}
\cmidrule(lr){8-10}
\cmidrule(lr){11-13}

& Win & Lose & Tie
& Win & Lose & Tie
& Win & Lose & Tie
& Win & Lose & Tie \\

\midrule

\textbf{Qwen-Image-Edit-Plus}
& \cellcolor{gray!15}58.51 & 39.36 & 2.13
& \cellcolor{gray!15}65.96 & 34.04 & 0.00
& \cellcolor{gray!15}63.83 & 29.79 & 6.38
& \cellcolor{gray!15}59.57 & 39.36 & 1.06 \\

\textbf{Nano Banana Pro}
& \cellcolor{gray!15}57.14 & 39.56 & 3.30
& \cellcolor{gray!15}55.68 & 44.32 & 0.00
& \cellcolor{gray!15}60.23 & 38.64 & 1.14
& \cellcolor{gray!15}61.36 & 37.50 & 1.14 \\

\textbf{Flux-2-Pro}
& \cellcolor{gray!15}60.23 & 39.77 & 0.00
& \cellcolor{gray!15}54.95 & 45.05 & 0.00
& \cellcolor{gray!15}59.34 & 37.36 & 3.30
& \cellcolor{gray!15}57.17 & 39.45 & 3.38 \\

\textbf{Seedream-4.5}
& 41.70 & \cellcolor{gray!15}56.04 & 2.26
& \cellcolor{gray!15}57.14 & 41.76 & 1.10
& \cellcolor{gray!15}49.45 & 44.21 & 6.34
& \cellcolor{gray!15}53.75 & 42.96 & 4.30 \\

\bottomrule
\end{tabular}
}
\end{table*}

\begin{table}[!htbp]
\centering
\caption{Average evaluation scores (1-10 scale) of CAMEO and direct editing baselines under four vision-language judges on the road anomaly insertion task. CAMEO consistently achieves higher scores across all judges.}
\label{tab:avg_scores_anomaly}

{\normalsize
\setlength{\tabcolsep}{3pt}
\begin{tabular}{l|cc|cc|cc|cc}
\toprule

\diagbox{\textbf{Backbone}}{\textbf{Judge}}
& \multicolumn{2}{c|}{\textbf{Qwen3-VL-Plus}}
& \multicolumn{2}{c|}{\textbf{GPT-4o}}
& \multicolumn{2}{c|}{\textbf{Gemini-2.5}}
& \multicolumn{2}{c}{\textbf{Claude-Opus-4.5}} \\

\cmidrule(lr){2-3}
\cmidrule(lr){4-5}
\cmidrule(lr){6-7}
\cmidrule(lr){8-9}

& CAMEO & Base
& CAMEO & Base
& CAMEO & Base
& CAMEO & Base \\

\midrule

\textbf{Qwen-Image-Edit-Plus}
& \cellcolor{gray!15}8.35 & 7.75
& \cellcolor{gray!15}8.00 & 7.65
& \cellcolor{gray!15}8.05 & 7.35
& \cellcolor{gray!15}8.30 & 7.50 \\

\textbf{Nano Banana Pro}
& \cellcolor{gray!15}8.28 & 7.30
& \cellcolor{gray!15}7.90 & 7.72
& \cellcolor{gray!15}8.45 & 7.42
& \cellcolor{gray!15}8.25 & 7.55 \\

\textbf{Flux-2-Pro}
& \cellcolor{gray!15}8.30 & 7.10
& \cellcolor{gray!15}8.05 & 7.75
& \cellcolor{gray!15}8.40 & 7.48
& \cellcolor{gray!15}8.28 & 7.25 \\

\textbf{Seedream-4.5}
& \cellcolor{gray!15}8.25 & 7.85
& \cellcolor{gray!15}7.95 & 7.60
& \cellcolor{gray!15}8.35 & 7.52
& \cellcolor{gray!15}8.22 & 7.60 \\

\bottomrule
\end{tabular}
}
\end{table}

\begin{table}[!htbp]
\centering
\caption{Average evaluation scores (1-10 scale) of CAMEO and direct editing baselines under four vision-language judges on the human pose switching task. CAMEO consistently achieves higher scores across most judges.}
\label{tab:avg_scores_pose}

{\normalsize
\setlength{\tabcolsep}{3pt}
\begin{tabular}{l|cc|cc|cc|cc}
\toprule

\diagbox{\textbf{Backbone}}{\textbf{Judge}}
& \multicolumn{2}{c|}{\textbf{Qwen3-VL-Plus}}
& \multicolumn{2}{c|}{\textbf{GPT-4o}}
& \multicolumn{2}{c|}{\textbf{Gemini-2.5}}
& \multicolumn{2}{c}{\textbf{Claude-Opus-4.5}} \\

\cmidrule(lr){2-3}
\cmidrule(lr){4-5}
\cmidrule(lr){6-7}
\cmidrule(lr){8-9}

& CAMEO & Base
& CAMEO & Base
& CAMEO & Base
& CAMEO & Base \\

\midrule

\textbf{Qwen-Image-Edit-Plus}
& \cellcolor{gray!15}8.28 & 7.67
& \cellcolor{gray!15}7.84 & 6.94
& \cellcolor{gray!15}7.88 & 6.34
& \cellcolor{gray!15}7.05 & 6.79\\

\textbf{Nano Banana Pro}
& \cellcolor{gray!15}8.28 & 7.80
& \cellcolor{gray!15}7.67 & 7.14
& \cellcolor{gray!15}7.82 & 6.62
& \cellcolor{gray!15}7.09 & 7.04\\

\textbf{Flux-2-Pro}
& \cellcolor{gray!15}8.47 & 7.73
& \cellcolor{gray!15}7.68 & 7.32
& \cellcolor{gray!15}7.50 & 6.32
& \cellcolor{gray!15}6.91 & 6.82 \\

\textbf{Seedream-4.5}
& 8.08 & \cellcolor{gray!15}8.46
& \cellcolor{gray!15}7.76 & 7.56
& \cellcolor{gray!15}7.29 & 7.19
& \cellcolor{gray!15}7.15 & 7.14 \\

\bottomrule
\end{tabular}
}
\end{table}

\subsection{Ablation Study}

To analyze the contribution of key components in CAMEO, we perform ablation experiments by selectively removing individual mechanisms while keeping the backbone models and evaluation protocols unchanged. Detailed quantitative results and experimental setup are reported in the Appendix.

\noindent\textbf{w/o Adaptive Reference Grounding.}
We disable adaptive reference grounding and perform editing using prompts only.

\noindent\textbf{w/o Quality Control.}
We remove the Quality Critic and Refinement Editor, reducing CAMEO to a structured single-pass generation pipeline.

\noindent\textbf{w/o Iterative Refinement.}
We retain evaluation but disable iterative correction, using only the first generated hypothesis.

Across both road anomaly insertion and human pose switching tasks, removing any component consistently degrades performance. In particular, adaptive reference grounding provides useful structural priors for complex transformations, quality control helps detect structural and contextual inconsistencies, and iterative refinement stabilizes editing under multiple constraints. The full CAMEO system achieves the best results, demonstrating the effectiveness of hierarchical coordination and closed-loop editing.

\subsection{Human Evaluation}

In addition to automated evaluation by vision-language judges, we conduct human preference studies to assess perceptual realism and instruction adherence.
For each task, we randomly sample edited image pairs generated by CAMEO and direct editing baselines.
Human annotators select the preferred result in each pairwise comparison. The detailed annotation protocol is provided in the Appendix. Table~\ref{tab:human_arena} reports aggregated human preference statistics. Across both road anomaly insertion and human pose switching tasks, CAMEO is consistently preferred over direct editing baselines. These preference trends are consistent with the automated evaluation results, indicating that structured coordination improves perceptual quality and multi-constraint consistency under human judgment.

\begin{table}[!htbp]
\centering
\caption{Human evaluation results (\%) comparing CAMEO with direct editing baselines.}
\label{tab:human_arena}

{\normalsize
\begin{tabular}{l|cccc}
\toprule
\textbf{Baseline} & \textbf{Win} & \textbf{Lose} & \textbf{Tie} & \textbf{Net $\uparrow$} \\
\midrule
Qwen-Image-Edit-Plus & 56.4 & 22.1 & 21.5 & \textbf{+34.3} \\
Nano Banana Pro      & 39.8 & 28.6 & 31.6 & \textbf{+11.2} \\
Flux-2-Pro           & 47.5 & 24.3 & 28.2 & \textbf{+23.2} \\
Seedream-4.5         & 33.7 & 29.4 & 36.9 & \textbf{+4.3} \\
\bottomrule
\end{tabular}
}
\end{table}

\subsection{Human Pose Switching Benchmark}

Since existing image editing benchmarks rarely provide high-quality evaluation settings specifically for human pose switching, we construct a dedicated benchmark for this task. Each sample consists of four elements: an original image, a pose switching instruction, a reference image specifying the target pose, and the resulting edited image. The original images are collected from the Pexels platform via API and exhibit diverse characteristics, including varied genders, ethnic groups, backgrounds, and camera orientations. Pose switching instructions are sampled from a predefined set of 30 manually designed pose transformations covering common full-body movements and viewpoint variations. This design introduces diverse structural transformations for evaluating pose consistency and anatomical plausibility in edited results. Further details of the benchmark are provided in the Appendix. The benchmark will be publicly released upon acceptance.

\section{Conclusion}

We revisit conditional image editing from the perspective of multi-constraint consistency and observe that most existing approaches follow an open-loop formulation, where semantic, structural, and contextual constraints are implicitly satisfied within a single generative pass. While effective for moderate edits, this design becomes unreliable as transformation complexity increases. We propose CAMEO, a hierarchical multi-agent framework that reformulates conditional editing as a structured, feedback-driven process. Through task-adaptive constraint activation, adaptive reference grounding, and quality aware controlling within an iterative refinement loop, CAMEO progressively regulates constraint satisfaction rather than relying on one-shot generation. Experiments on road anomaly insertion and human pose switching demonstrate improved robustness and multi-constraint consistency across multiple editing backbones and evaluation models.

\noindent\textbf{Limitations and Future Work.} CAMEO introduces additional computational overhead due to iterative coordination and remains limited by the capabilities of the underlying editing and evaluation models. Future work includes developing more reliable evaluation metrics for measuring image editing quality.

\clearpage
%\section*{Acknowledgements}
%Please insert your acknowledgments here.

\bibliographystyle{splncs04}
%\bibliography{main}
\bibliography{main}

\newpage
\appendix
\section*{Appendix Overview}
This appendix provides additional details and supporting materials for the proposed framework. Section~A describes the model configuration of each agent in the multi-agent system. Section~B introduces the evaluation protocol used in our experiments, while Section~C presents the ablation study analyzing the contribution of key components. 
Section~D details the human evaluation protocol, and Section~E provides a running case illustrating the full workflow of CAMEO.

\section{Model Configuration of Each Agent}
In this section, we describe the underlying models used to implement each component of the proposed multi-agent workflow. Each agent in the pipeline is instantiated with a specific model according to its functional role.

\noindent\textbf{Strategic Director.}
The Strategic Director is responsible for interpreting the editing instruction and determining the overall task strategy. This agent is implemented using Qwen3-VL-Plus, which provides strong multimodal reasoning capabilities for understanding both textual instructions and visual context.

\noindent\textbf{Visual Research Specialist.}
The Visual Research Specialist is responsible for retrieving or synthesizing reference images that provide additional visual guidance for the editing task. For reference retrieval, we employ a web image search API. Given the textual description of the desired reference, the system first queries the Serp API to obtain a set of candidate images. The Visual Research Specialist then analyzes the retrieved results and selects the most relevant references according to the editing instruction and visual context. For reference image selection, we employ Gemini 2.5 Flash as the underlying vision–language model. For reference synthesis, when suitable images cannot be directly retrieved, we generate reference images using Nano Banana Pro. In this case, the model produces a reference image that satisfies the required semantic attributes described in the instruction, which is subsequently used to guide the editing process.

\noindent\textbf{Instruction Architect.}
The Instruction Architect transforms high-level editing instructions into structured prompts suitable for image editing models. This agent is implemented using GPT-4o to generate constraint-aware editing prompts.

\noindent\textbf{Generative Creator.}
The Generative Creator performs the actual image editing process. In our experiments, we use Nano Banana Pro as the primary image editing model to generate the edited image according to the structured prompt, potential reference image, and source image.

\noindent\textbf{Quality Critic.}
The Quality Critic evaluates the generated images across multiple dimensions. This component is implemented using Qwen3-VL-Plus, which provides strong multimodal reasoning capability for fine-grained visual quality assessment.

\noindent\textbf{Refinement Editor.}
The Refinement Editor iteratively improves the generated result based on the feedback provided by the Quality Critic. 
At each refinement step, the diagnostic feedback is translated into an updated editing instruction, and the image is re-edited accordingly.  This component is implemented using Qwen Image Edit Plus to perform successive editing iterations, enabling progressive improvement of semantic alignment and visual realism. We select Qwen Image Edit Plus partly due to its strong capability in text-aware image editing, which is beneficial for tasks that require modifying textual elements within the scene such as Chinese context.

\noindent\textbf{Modularity of Agent Design.}
The proposed multi-agent workflow is modular, where each agent functions as an independent component with clearly defined responsibilities. As a result, individual agents can be replaced with alternative models that provide similar capabilities. The models described above correspond to the configuration used in our experiments, but the overall framework is not tied to any specific model and can be extended with other vision-language models, retrieval systems, or image editing models.
\section{Evaluation Protocol}

\subsection{Overview}

We employ vision-language judges as automated judges to assess the quality of edited images.
For each evaluation case, the judge model is provided with the editing instruction and two edited images generated by different methods. Following an arena-style evaluation protocol, the vision-language judge performs pairwise comparisons between candidate images and determines which result better satisfies the editing instruction while maintaining visual plausibility and overall visual quality. Each comparison involves two images produced by different methods (e.g., our method versus a baseline model). To mitigate potential position bias, we adopt a counterbalanced ordering strategy. Specifically, in odd-numbered comparison cases the image generated by our method is presented first, while in even-numbered cases the baseline image is presented first. Since the two tasks considered in this work, road anomaly insertion and human pose switching, exhibit different visual characteristics and evaluation requirements, we adopt task-specific prompt templates for the evaluation models. The evaluation criteria remain conceptually consistent, but the prompt descriptions are adapted to better reflect the visual properties of each task.

\subsection{Prompt for Road Anomaly Insertion Evaluation}

For the road anomaly insertion task, the evaluation focuses on whether the generated image correctly reflects the intended anomaly insertion and environmental changes while maintaining realistic physical properties and consistency with the surrounding road scene. During evaluation, the vision-language judge is provided with the editing instruction and two candidate images generated by different methods (e.g., our method and a baseline). The judge model performs pairwise comparison in an arena-style setting and determines which image better satisfies the editing instruction while maintaining realistic anomaly appearance and overall visual plausibility.

The following template of prompts is used for vision-language judges.

\begin{tcolorbox}[
colback=black!3,
colframe=black!40,
breakable
]
{\ttfamily\footnotesize

You are a professional visual quality reviewer.

Two images (A and B) are generated from the SAME original road image using the SAME editing instruction.

Known conditions:\\
- Inserted anomalies: <anomaly\_types>\\
- Weather change: <weather\_condition>

Evaluation dimensions and scoring guidelines (1--10 scale):

For each dimension below, assign an integer score from 1 to 10.

Evaluation dimensions:

1. Semantic correctness\\
How well the inserted anomalies and weather change match the intended descriptions.

2. Physical plausibility\\
Whether the anomalies and environmental changes obey basic physical laws (e.g., gravity, contact, scale, geometry, lighting consistency under the specified weather).

3. Boundary blending\\
How naturally the anomalies blend with the surrounding road surface (e.g., edges, texture, lighting, transition with nearby regions).

4. Contextual coherence\\
How well the anomalies and weather conditions fit into the overall scene (e.g., traffic setting, road type, weather consistency, time of day).

Scoring rules:\\
- Use only integer scores from 1 to 10.\\
- Be strict and comparative between A and B.\\
- If both images perform similarly across all dimensions, you may declare a tie.

In addition to dimension scores, provide an overall\_score (1--10) for each image that reflects its holistic quality across all four dimensions.

Output STRICTLY the following JSON and nothing else:

\{
  "scores": \{
    "A": \{
      "semantic": int,
      "physical": int,
      "blending": int,
      "context": int,
      "overall\_score": int
    \},
    "B": \{
      "semantic": int,
      "physical": int,
      "blending": int,
      "context": int,
      "overall\_score": int
    \}
  \},
  "winner": "A|B|tie",
  "reason": "brief explanation"
\}

}
\end{tcolorbox}

\subsection{Prompt for Human Pose Switching Evaluation}

For the human pose switching task, the evaluation focuses on whether the generated image correctly follows the target pose instruction while maintaining realistic human body structure and visual consistency with the scene. During evaluation, the vision-language judge is provided with the target pose instruction and two candidate images generated by different methods (e.g., our method and a baseline). The judge model performs pairwise comparison in an arena-style setting and determines which image better satisfies the pose instruction while maintaining realistic body structure and overall visual quality.

The following template of prompts is used for vision-language judges.

\begin{tcolorbox}[
colback=black!3,
colframe=black!40,
breakable
]
{\ttfamily\footnotesize

You are a professional visual quality reviewer.

Two images (A and B) are generated using the SAME target pose instruction.

Known conditions:\\
- Target pose instruction: <pose\_instruction>

Evaluation dimensions and scoring guidelines (1--10 scale):

1. Semantic correctness\\
How accurately the generated human pose matches the target pose instruction, including limb orientation, joint configuration, and overall body posture.

2. Physical plausibility\\
Whether the human body follows realistic anatomical structure and joint articulation (e.g., limb proportions, joint angles, natural posture).

3. Boundary blending\\
How naturally the edited human figure blends with the surrounding scene (e.g., edges, lighting, shadows, interaction with the background).

4. Contextual coherence\\
How well the edited person fits into the overall scene (e.g., body orientation, interaction with objects, spatial consistency).

Scoring rules:\\
- Use only integer scores from 1 to 10.\\
- Be strict and comparative between A and B.\\
- If both images perform similarly across all dimensions, you may declare a tie.

In addition to dimension scores, provide an overall\_score (1--10) for each image that reflects its holistic quality across all four dimensions.

Output STRICTLY the following JSON and nothing else:

\{
  "scores": \{
    "A": \{
      "semantic": int,
      "physical": int,
      "blending": int,
      "context": int,
      "overall\_score": int
    \},
    "B": \{
      "semantic": int,
      "physical": int,
      "blending": int,
      "context": int,
      "overall\_score": int
    \}
  \},
  "winner": "A|B|tie",
  "reason": "brief explanation"
\}

}
\end{tcolorbox}

\section{Ablation Study}

To better understand the contribution of each component in our multi-agent framework, we conduct an ablation study by selectively removing or modifying key modules in the pipeline. In particular, we analyze the impact of several design choices, including the reference image retrieval mechanism, the quality control module, and the iterative refinement process. For each ablation setting, we keep all other components unchanged and evaluate the resulting image editing performance using the same evaluation protocol described in Sec.~B. The ablation experiments are conducted on a subset of 500 images for each task for efficiency. The evaluation is performed using the same vision-language judges adopted in the main experiments. The quantitative results are illustrated in the following figures, where each plot compares the performance of our full system with its ablated variants.

\noindent\textbf{w/o Adaptive Reference Grounding.}
We disable adaptive reference grounding and perform editing using prompts only, removing the reference retrieval and synthesis stage from the pipeline. 

\noindent\textbf{w/o Quality Control.}
We remove the Quality Critic and Refinement Editor, reducing CAMEO to a structured single-pass generation pipeline without evaluation or correction. 

\noindent\textbf{w/o Iterative Refinement.}
We retain the Quality Critic but disable iterative correction, using only the first generated hypothesis without refinement.

\begin{figure}[!htbp]
\centering
\includegraphics[width=\linewidth]{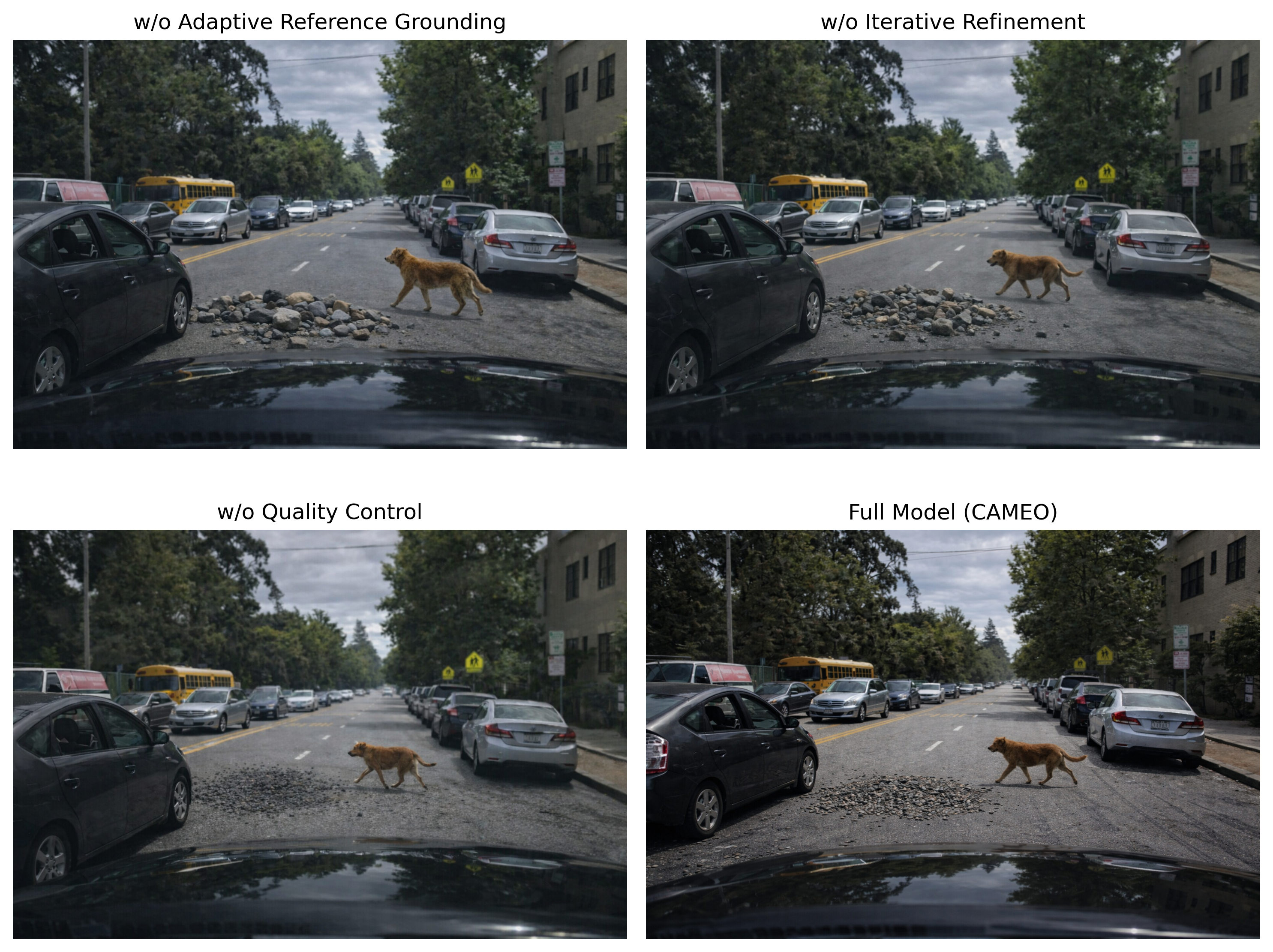}
\caption{Qualitative comparison of the full CAMEO system and three ablation variants. Removing key components leads to degraded realism and quality of details.}
\label{fig:ablation_visual}
\end{figure}

Figure~\ref{fig:ablation_visual} shows qualitative comparisons between the full CAMEO model and different ablation settings. The quantitative comparison between the full model and its ablated variants is summarized in Table~\ref{tab:anomaly_ablation} and Table~\ref{tab:pose_ablation}.

\begin{table}[!htbp]
\centering
\caption{Ablation study of CAMEO on the road anomaly insertion task. Average evaluation scores (1--10 scale) reported by four vision-language judges.}
\label{tab:anomaly_ablation}
\resizebox{\linewidth}{!}{
\begin{tabular}{l|cccc}
\hline
Ablation Setting & Qwen3-VL-Plus & GPT-4o & Gemini-2.5 & Claude-Opus-4.5 \\
\hline
\textbf{Full Model (CAMEO)} & \textbf{8.47} & \textbf{8.42} & \textbf{8.36} & \textbf{8.45} \\
w/o Adaptive Reference Grounding & 7.88 & 7.91 & 7.80 & 7.83 \\
w/o Quality Control & 7.05 & 7.18 & 7.10 & 6.96 \\
w/o Iterative Refinement & 7.96 & 7.99 & 7.92 & 7.94 \\
\hline
\end{tabular}}
\end{table}

\begin{table}[!htbp]
\centering
\caption{Ablation study of CAMEO on the human pose switching task. Average evaluation scores (1--10 scale) reported by four vision-language judges.}
\label{tab:pose_ablation}
\resizebox{\linewidth}{!}{
\begin{tabular}{l|cccc}
\hline
Ablation Setting & Qwen3-VL-Plus & GPT-4o & Gemini-2.5 & Claude-Opus-4.5 \\
\hline
\textbf{Full Model (CAMEO)} & \textbf{8.39} & \textbf{8.33} & \textbf{8.27} & \textbf{8.35} \\
w/o Adaptive Reference Grounding & 7.83 & 7.86 & 7.74 & 7.79 \\
w/o Quality Control & 6.98 & 7.12 & 7.05 & 6.91 \\
w/o Iterative Refinement & 7.91 & 7.88 & 7.82 & 7.86 \\
\hline
\end{tabular}}
\end{table}

\section{Human Evaluation Protocol}

\begin{figure}[!htbp]
\centering
\includegraphics[width=\linewidth]{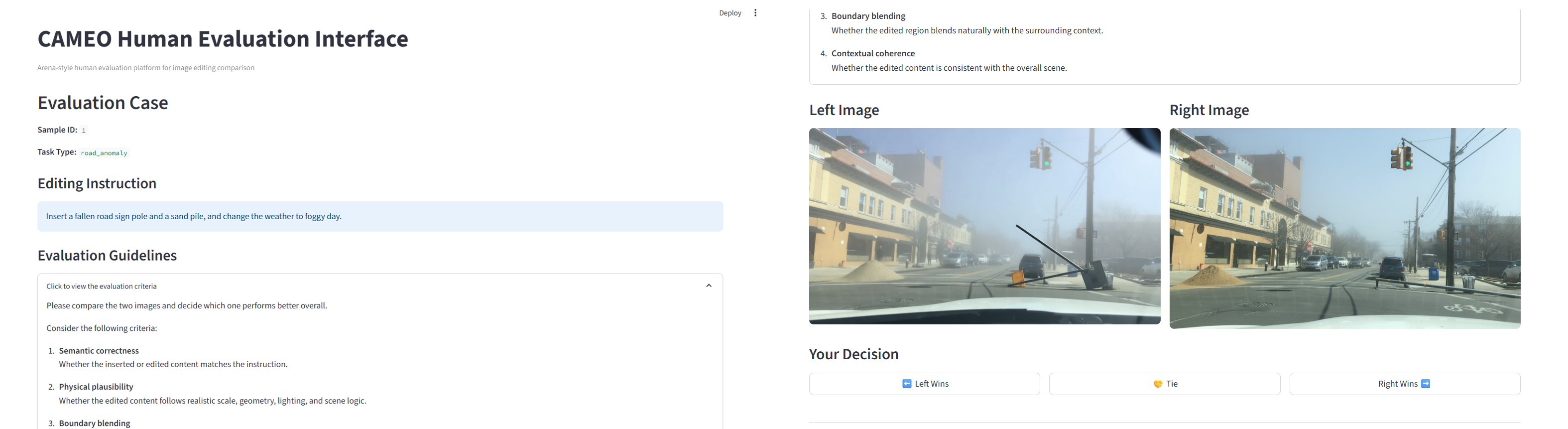}
\caption{Screenshot of the human evaluation interface used in our study.}
\label{fig:human_interface}
\end{figure}

In addition to automated evaluation using vision-language judges, we conduct a human evaluation to further assess the quality of the edited images. Similar to the automated evaluation protocol, human assessment follows an arena-style pairwise comparison setting. For each evaluation case, annotators are presented with the editing instruction together with two edited images generated by different methods (e.g., our method and a baseline). Annotators are asked to determine which image better satisfies the editing instruction while maintaining realistic visual appearance. We recruit 10 human annotators for this study, and each annotator evaluates 100 randomly sampled pairs. To facilitate the evaluation process, we develop a lightweight web-based interface for human annotators. As shown in Figure~\ref{fig:human_interface}, the interface displays the editing instruction and the two candidate images side by side, allowing annotators to directly compare the results and select the better image or declare a tie. To mitigate potential position bias, the order of the two candidate images is counterbalanced across evaluation cases. Specifically, in odd-numbered comparison cases the image generated by our method is presented first, while in even-numbered cases the baseline image is presented first. Annotators are blind to the identity of the compared methods. Unlike the automated evaluation, human annotators are not required to assign numerical scores. Instead, they compare the images while considering the same evaluation criteria used in the vision-language judge assessment, including semantic correctness, physical plausibility, boundary blending, and contextual coherence. For each comparison, annotators select the image that performs better overall according to these criteria, or declare a tie if the two results appear comparable. The final results are summarized as A/B/tie statistics across all evaluation cases.

\section{Running Case of CAMEO}

To provide a clearer understanding of the internal workflow of CAMEO, we present a detailed running example that illustrates the intermediate outputs produced by each agent in the pipeline. Starting from a given editing instruction, the system sequentially executes multiple agents, including task interpretation, reference retrieval or synthesis, prompt construction, image generation, quality assessment, and iterative refinement. For each stage, we show the corresponding intermediate outputs to demonstrate how the system progressively transforms the initial instruction into the final edited image. This example aims to provide a transparent view of the decision-making and generation process within the multi-agent framework, highlighting how different agents collaborate to achieve the final editing result.

%Strategic Director
\begin{tcolorbox}[
title=Strategic Director,
colback=blue!3,
colframe=blue!50!black,
fonttitle=\bfseries,
breakable
]

\textbf{Input Instruction}

Insert two road anomalies and modify the weather condition according to the instruction.

Example:
"Insert a pothole and a road crack on the road surface and change the weather to rainy."

\medskip

\textbf{Agent Reasoning}

The instruction contains three editing objectives:
\begin{itemize}
\item Anomaly insertion (two anomalies)
\item Environmental modification (weather change)
\end{itemize}

The system determines that visual references may help guide anomaly appearance and environmental effects.

\medskip

\textbf{Output}

\begin{tcolorbox}[
colback=black!3,
colframe=black!40,
breakable
]
{\ttfamily\footnotesize
\{
  "reference\_type": "image",
  "insertion\_content": ["pothole", "road\_crack"],
  "prompt\_guidance": "Refine an image editing prompt to insert the pothole and road crack, and modify the weather to rainy. Ensure the prompt clearly describes realistic placement, detailed scale, lighting, and scene consistency for the edited image.",

  "evaluation\_criteria": [
    "semantic correctness",
    "physical plausibility",
    "boundary blending",
    "contextual coherence"
  ]
\}
}
\end{tcolorbox}

\end{tcolorbox}

%Visual Research Specialist
\begin{tcolorbox}[
title=Visual Research Specialist,
colback=green!3,
colframe=green!50!black,
fonttitle=\bfseries,
breakable
]

\textbf{Input from Strategic Director}

Reference request:
\begin{itemize}
\item anomaly types: pothole, road crack
\item weather condition: rainy
\item reference type: image
\end{itemize}

\medskip

\textbf{Agent Reasoning}

The agent attempts to obtain visual references that match the requested insertion contents and environmental conditions.

Two strategies are considered:

\begin{itemize}
\item \textbf{Reference retrieval:} search the web for real images containing the target anomalies and weather condition.
\item \textbf{Reference synthesis:} if suitable references cannot be retrieved, generate a reference image using an image generation model.
\end{itemize}

\medskip

\textbf{Reference Retrieval}

\textbf{Search tool:} Web search via SerpAPI

\textbf{Search query example:}
\begin{tcolorbox}[colback=black!3,colframe=black!40]
\texttt{pothole under rainy day; road crack under rainy day}
\end{tcolorbox}
\textbf{Candidate images retrieved:}

(top search results from web image search)

\textbf{Filtering step:}

A vision-language model evaluates candidate images and selects the most relevant reference according to visual realism and consistency with the target insertion context. Then it store selected images for each insertion content as reference images. Here is an example prompt sent:
\begin{tcolorbox}[
colback=black!3,
colframe=black!40,
breakable
]
{\ttfamily\footnotesize

You are an image quality inspection assistant. A set of images is provided as candidate reference images.

Your task is to evaluate each image and select only the images that pass all filtering criteria.

For each image, evaluate the following four criteria:

1. Real-world photograph  
Determine whether the image is a real-world photograph rather than an illustration, rendering, cartoon, 3D scene, or game screenshot.

2. Not AI-generated  
If the image shows obvious artifacts typical of AI-generated images (e.g., unnatural textures, distorted objects, inconsistent structures, strange text), mark it as AI-generated.

3. No watermark  
If the image contains visible copyright watermarks or noticeable logos/text overlays, mark it as having a watermark.

4. Matches the content  
Determine whether the image clearly contains the specified pothole under rainy day and whether the it is the primary subject rather than being distant, blurry, or unrelated.

The target content is:

pothole under rainy day

Output strictly the following JSON format and nothing else:

\{
  "accepted\_indices": [array of integers starting from 0],
  "detail": [
    \{
      "index": 0,
      "realistic": true,
      "ai\_generated": false,
      "has\_watermark": false,
      "matches": true,
      "comment": "one short explanation"
    \}
  ]
\}

If an image fails any of the criteria (not realistic, suspected AI-generated, contains watermark, or does not match the content), its index should NOT be included in \texttt{accepted\_indices}.

}
\end{tcolorbox}

\medskip

\textbf{Alternative Strategy: Reference Synthesis}

If suitable web references are unavailable, the agent generates a reference image using an image generation model conditioned on the requested insertion content

\medskip

\textbf{Output}

Selected reference image that visually represents the requested insertion contents. The reference is then passed to the Generative Creator.

\end{tcolorbox}

%Prompt Architect

\begin{tcolorbox}[
title=Prompt Architect,
colback=orange!3,
colframe=orange!60!black,
fonttitle=\bfseries,
breakable
]

\textbf{Input from Strategic Director}

\begin{itemize}
\item Editing instruction from the user
\item Task plan from the Strategic Director
\item Target anomalies: pothole, road crack
\item Weather condition: rainy
\end{itemize}

\medskip

\textbf{Agent Reasoning}

The agent constructs a structured editing prompt that clearly specifies the required modifications.  
The prompt must include the target anomalies, environmental changes, and contextual constraints to ensure realistic generation.

Key considerations:
\begin{itemize}
\item explicitly describe the anomaly types to be inserted
\item specify the weather modification
\item specify visual attributes of the inserted anomalies (e.g., color, texture, and shape)
\item describe geometric and spatial attributes such as size, placement location, and relative distance within the scene
\item ensure the edits are consistent with the surrounding scene
\item maintain realistic road geometry and lighting
\end{itemize}

\medskip

\textbf{Example Prompt Output}
\begin{tcolorbox}[colback=black!3,colframe=black!40]
\texttt{
Insert two road anomalies into the scene: a pothole and a road crack on the road surface. The pothole should appear in the center lane with realistic depth, rough edges, and dark interior shading.
The road crack should extend naturally along the road surface with irregular shape and texture. Change the weather condition to rainy. The road surface should appear wet with subtle reflections and slightly darker pavement color. Ensure the anomalies follow realistic geometry and scale, blend naturally with the surrounding road texture, and remain consistent with the lighting and perspective of the scene.
}
\end{tcolorbox}
\end{tcolorbox}

%Generative Creator
\begin{tcolorbox}[
title=Generative Creator,
colback=purple!3,
colframe=purple!60!black,
fonttitle=\bfseries,
breakable
]

\textbf{Input}

\begin{itemize}
\item Original input road image
\item Structured editing prompt from the Prompt Architect
\item (Optional) reference image from the Visual Research Specialist
\end{itemize}

\medskip

\textbf{Agent Reasoning}

The agent performs image editing according to the structured prompt and optional reference guidance.
The editing model inserts the specified anomalies and modifies the environmental condition while preserving the original scene structure.

Key objectives include:

\begin{itemize}
\item inserting the target anomalies at realistic locations
\item applying the requested weather modification
\item strictly follow the instruction
\end{itemize}

\medskip

\textbf{Output}

An edited image containing the inserted anomalies and modified weather condition.

\medskip

%\textbf{Example Result}

%\begin{center}
%\includegraphics[width=0.6\linewidth]{example_generated_image.png}
%\end{center}

\end{tcolorbox}

%Quality Critic

\begin{tcolorbox}[
title=Quality Critic,
colback=red!3,
colframe=red!60!black,
fonttitle=\bfseries,
breakable
]

\textbf{Input}

\begin{itemize}
\item Edited image produced by the Generative Creator
\item Editing instruction
\item Evaluation dimensions provided by the Strategic Director
\item Target insertion contents (anomalies) identified by the Strategic Director
\end{itemize}

\medskip

\textbf{Agent Reasoning}

The evaluation criteria are determined by the Strategic Director according to the task requirements. Based on the selected evaluation dimensions, the Quality Critic evaluates each inserted anomaly independently rather than scoring the entire image as a whole.

For the road anomaly insertion task, the selected evaluation dimensions include:

\begin{itemize}
\item semantic correctness
\item physical plausibility
\item boundary blending
\item context coherence
\end{itemize}

For each insertion content (i.e., each anomaly), the agent assigns scores on a scale of 1--5 for all evaluation dimensions and produces structured diagnostic feedback.

If any evaluation dimension of any anomaly falls below the predefined threshold (score $<3$), the result is considered unsatisfactory and the image is sent to the Refinement Editor together with a fix comment. Otherwise, the result is accepted and the generation process terminates.

\medskip

\textbf{Example 1: Failed Evaluation (Refinement Required)}

\begin{tcolorbox}[colback=black!3,colframe=black!40]

\texttt{
\{
  "insertion\_1": \{
    "type": "pothole",
    "semantic": 4,
    "physical": 3,
    "blending": 2,
    "context": 3
  \},
  "insertion\_2": \{
    "type": "road\_crack",
    "semantic": 4,
    "physical": 4,
    "blending": 3,
    "context": 3
  \},
  "decision": "refine",
  "fix\_comment": "The pothole boundary blending with the surrounding road surface is unnatural. The transition between the anomaly and the pavement texture should be smoother."
\}
}

\end{tcolorbox}

\medskip

Since the boundary blending score of anomaly\_1 falls below the threshold, the system triggers the refinement stage.

\medskip

\textbf{Example 2: Successful Evaluation (Accepted)}

\begin{tcolorbox}[colback=black!3,colframe=black!40]

\texttt{
\{
  "insertion\_1": \{
    "type": "pothole",
    "semantic": 4,
    "physical": 4,
    "blending": 3,
    "context": 4
  \},
  "insertion\_2": \{
    "type": "road\_crack",
    "semantic": 4,
    "physical": 3,
    "blending": 3,
    "context": 4
  \},
  "decision": "pass",
  "fix\_comment": ""
\}
}

\end{tcolorbox}

\medskip

The diagnostic feedback is then passed to the Refinement Editor to guide the next editing iteration.

\end{tcolorbox}

%Refinement Editor
\begin{tcolorbox}[
title=Refinement Editor,
colback=cyan!3,
colframe=cyan!60!black,
fonttitle=\bfseries,
breakable
]

\textbf{Input}

\begin{itemize}
\item Edited image from the Generative Creator
\item Evaluation results from the Quality Critic
\item Fix comment describing the detected issue
\end{itemize}

\medskip

\textbf{Agent Reasoning}

If the Quality Critic determines that an insertion content fails the evaluation threshold, a fix comment describing the detected issue is produced.  
The system directly uses this fix comment as the modification instruction for the Refinement Editor.

Since the Refinement Editor is an image editing model rather than a reasoning module, it does not generate new textual instructions. Instead, it receives the fix comment and applies the corresponding visual correction to the image.

\medskip

\textbf{Fix Comment from Quality Critic}

\begin{tcolorbox}[colback=black!3,colframe=black!40]
\ttfamily
"The pothole boundary blending with the surrounding road surface is unnatural. 
The transition between the anomaly and the pavement texture should be smoother."
\end{tcolorbox}

\medskip

\textbf{Updated Editing Prompt}

\begin{tcolorbox}[colback=black!3,colframe=black!40]
\ttfamily
Modify the edited image according to the following instruction:

"The pothole boundary blending with the surrounding road surface is unnatural. 
The transition between the anomaly and the pavement texture should be smoother."

DO NOT change other elements.
\end{tcolorbox}

\medskip

\textbf{Output}

A refined edited image generated by applying the fix instruction.

\medskip

\textbf{Iterative Refinement}

After the refinement step, the updated result is sent back to the Quality Critic for evaluation using the same criteria as before. If the refined image still fails to meet the threshold requirements, a new fix comment is generated and another refinement iteration is performed.

This refinement–evaluation loop continues until the image passes the evaluation or the maximum number of refinement attempts is reached. In our system, the refinement process is limited to three iterations. If the result still fails after three refinement rounds, the current editing task is discarded.

\end{tcolorbox}

\end{document}